\documentclass[conference]{IEEEtran}
\pdfoutput=1
\usepackage{times}

\usepackage[numbers,sort]{natbib}
\usepackage{multicol}

\usepackage{epsfig}

\usepackage[crop=pdfcrop,mode=batch]{pstool}

\usepackage{epstopdf}

\usepackage{graphicx}
\usepackage[draft]{hyperref}

\usepackage{flushend}

\usepackage{amsmath,amssymb,bm}

\usepackage[ruled,linesnumbered,vlined]{algorithm2e}

\usepackage{multirow}
\usepackage{booktabs}
\usepackage{arydshln}

\newcommand{\R}[1]{\ensuremath{\mathbb{R}^{#1}}}

\newcommand{\vr}[1]{{\mbox{\bm{$#1$}}}}

\newcommand{\mt}[1]{{\mbox{\bm{$#1$}}}}

\newcommand{\manif}[1]{\ensuremath{\mathcal{#1}}}

\newcommand{\C}{\ensuremath{\manif{C}}} 
\newcommand{\X}{\ensuremath{\manif{X}}} 
\newcommand{\Xfree}{\ensuremath{\manif{X}_{\text{free}}}} 
\newcommand{\U}{\ensuremath{\manif{U}}} 

\newcommand{\TX}[1]{\ensuremath{\mathcal{T}_{#1}\X}}

\newcommand{\tr}{^{\!\top}}
\newcommand{\inv}{^{-1}}

\graphicspath{{./}{figures/}}

\begin{document}

\title{Kinodynamic Planning on Constraint Manifolds}


\author{Ricard Bordalba, Llu\'{\i}s Ros, and Josep M. Porta \\ 
Institut de Rob\`otica i Inform\`atica
Industrial, CSIC-UPC,  Barcelona, Spain\\
E-mails: \{rbordalba,porta,ros\}@iri.upc.edu
}


\maketitle

\begin{abstract}
This paper presents a motion planner for systems subject to kinematic
and dynamic constraints. The former appear when kinematic loops are
present in the system, such as in parallel manipulators, in robots
that cooperate to achieve a given task, or in situations involving
contacts with the environment. The latter are necessary to obtain
realistic trajectories, taking into account the forces acting on the
system. The kinematic constraints make the state space become an
implicitly-defined manifold, which complicates the application of
common motion planning techniques. To address this issue, the planner
constructs an atlas of the state space manifold incrementally, and
uses this atlas both to generate random states and to dynamically
simulate the steering of the system towards such states. The resulting
tools are then exploited to construct a rapidly-exploring random tree
(RRT) over the state space. To the best of our knowledge, this is the
first randomized kinodynamic planner for implicitly-defined state
spaces. The test cases presented in this paper validate the approach
in significantly-complex systems.
\end{abstract}

\IEEEpeerreviewmaketitle

\section{Introduction} \label{sec:introduction}

The motion planning problem has been a subject of active research
since the early days of Robotics~\cite{Labombe_91}. Although it can be
formalized in simple terms---find a feasible trajectory to move a
robot between two states---and despite the significant advances in the
field, it is still an open problem in many respects. The complexity of
the problem arises from the multiple constraints that have to be taken
into account, such as potential collisions with static or moving
objects in the environment, kinematic loop-closure constraints, torque
and velocity limits, or energy and time execution bounds, to name a
few. All these constraints are relevant in the factory and home
environments in which Robotics is called to play a fundamental role in
the near future.

\begin{figure}[t]

  \begin{center}
   \pstool[width=.8\linewidth]{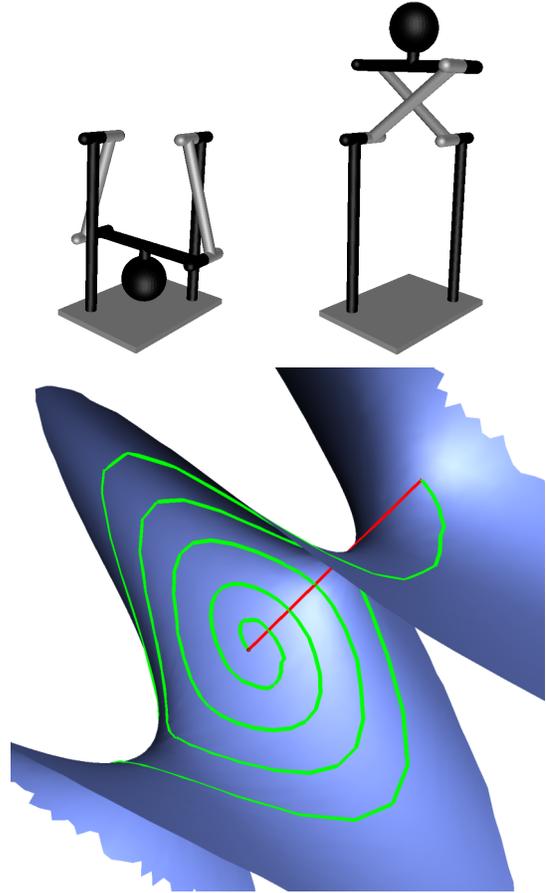}{
      \psfrag{a}[l]{}
      \psfrag{b}[l]{}	
      \psfrag{c}[l]{}	
    }
  \end{center}
  \vspace{-0.25cm}
  \caption{A kindoynamic planning problem on a four-bar pendulum
  modeling a swing boat ride. Top: The start and goal states, both with
  null velocity. Bottom:~The kinematic constraints define an
  helicoidal manifold. A kinematically-feasible trajectory (red) and a
  trajectory also fulfilling dynamic constraints (green) may be quite different.}
  \label{fig:pendulum_mech}
\end{figure}

The complexity of the problem is typically tackled by first relaxing
some of the constraints. For example, while obstacle avoidance is a
fundamental issue, the lazy approaches initially disregard
it~\cite{Bohlin_ICRA2000}. Other approaches concentrate on
geometric~\cite{Kavraki_TRA1996} and kinematic
feasibility~\cite{Jaillet_TRO2013} from the outset, which constitute
already challenging issues by themselves. In these and other
approaches~\cite{Dubowsky_ROMANSY1985}, dynamic constraints such as
speed, acceleration, or torque limits are neglected, with the hope
that they will be enforced in a postprocessing stage. Decoupled
approaches, however, may not lead to solutions satisfying all the
constraints. It is not difficult to find situations in which
a kinematically-feasible, collision-free trajectory becomes unusable
because it does not account for the system
dynamics~(Fig.~\ref{fig:pendulum_mech}).

\begin{figure}[t]
  \begin{center}
    \pstool[width=\linewidth]{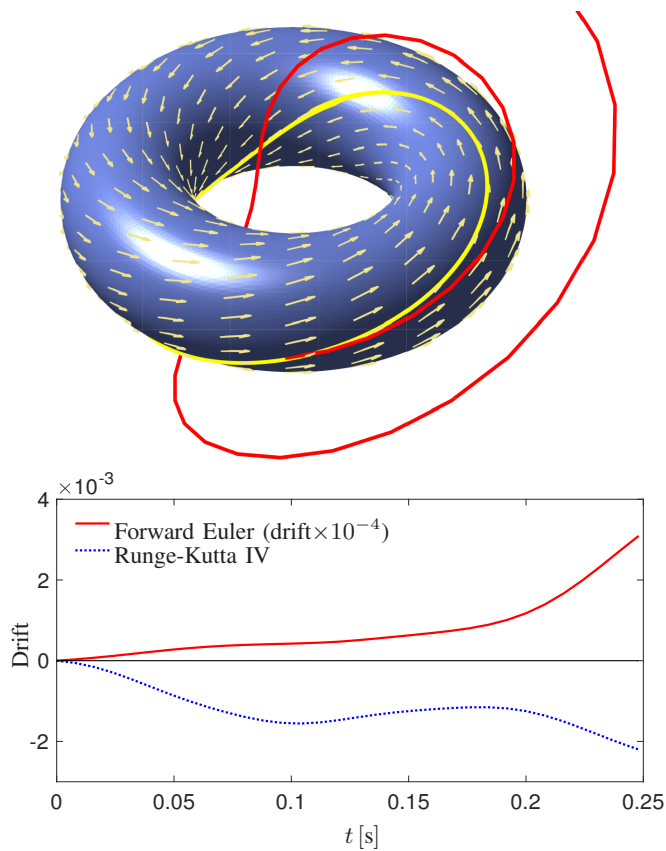}{
      \psfrag{a}[l]{(a)}
      \psfrag{b}[l]{(b)}	
      \psfrag{x}[l]{\small $t\,[\textrm{s}]$}
      \psfrag{y}[c]{\small \hspace{-0.1cm} Drift}	
      \psfrag{legeng1}[l]{\small Forward Euler (drift$\times 10^{-4}$)}	
      \psfrag{legend2}[l]{\small Runge-Kutta IV}	
    }
  \end{center}
\vspace{-0.25cm}
  \caption{Drift caused by numerical integration of an ODE system.
  Top:~A particle moving on a torus under the shown vector field. The
  trajectory obtained by the forward Euler method (red) increasingly
  diverges from the exact trajectory (yellow). Bottom: With more
  accurate procedures, such as the 4th order Runge-Kutta method
  (blue), the drift may be reduced, but not canceled.}
  \label{fig:integration}
\vspace{-0.2cm}
\end{figure}

This paper presents a sampling-based planner that simultaneously
considers collision avoidance, kinematic, and dynamic constraints. The
planner constructs a bidirectional rapidly-exploring random tree (RRT)
on the state space manifold implicitly defined by the kinematic
constraints. In the literature, the suggested way to define an RRT
including such constraints is to differentiate them and add them to
the ordinary differential equations (ODE) defined by the dynamic
constraints~\cite{Lavalle_06}. In such an approach, however, the
underlying geometry of the problem would be lost. The random samples
used to guide the RRT extension would not be generated on the state
space manifold, but in the larger ambient space, which results in
inefficiencies~\cite{Jaillet_TRO2013}. The numerical integration of
the resulting ODE system, moreover, would be affected by
drift~(Fig.~\ref{fig:integration}). In some applications, such a drift
might be tolerated, but in others, such as in robots with closed
kinematic loops, it would render the simulation unprofitable due to
unwanted link penetrations, disassemblies, or contact losses. In this
paper, the sampling and drift issues are addressed by preserving the
underlying geometry of the problem. To this end, we propose to combine
the extension of the RRT with the incremental construction of an atlas
of the state space manifold~\cite{Lee_2001}. The atlas is enlarged as
the RRT branches reach yet unexplored areas of the manifold. Moreover,
it is used to effectively generate random states and to dynamically
simulate the steering of the system towards such states.

This paper is organized as follows. Section~\ref{sec:related} puts the
proposed planner in the context of existing approaches.
Section~\ref{sec:formalization} formalizes the problem and paves the
way to Section~\ref{sec:state_space_atlas}, which describes the
fundamental tools to map and explore an implicitly-defined state
space. The resulting planner is described in
Section~\ref{sec:theplanner} and experimentally validated in
Section~\ref{sec:experiments}. Finally, Section~\ref{sec:conclusion}
concludes the paper and discusses points deserving further attention.

\section{Related Work} \label{sec:related}

The problem of planning under dynamic constraints, also known as
kinodynamic planning~\cite{Lavalle_06}, is harder than planning with
geometric constraints, which is already known to be
PSPACE-hard~\cite{Reif_ASFCS1979,Canny_STC1988}. Although particular,
exact solutions for some systems have been
given~\cite{Donald_JACM1993}, general solutions do also exist. Dynamic
programming approaches, for example, define a grid of cost-to-go
values to search for a
solution~\cite{Barraquand_ICRA1991,Lynch_IJRR1996,Cherif_ICRA1999},
and can compute accurate solutions in lower-dimensional problems. Such
an approach, however, does not scale well to problems with many
degrees of freedom. In contrast, numerical optimization
techniques~\cite{Kalakrishnan_2011,Posa_ICRA2016,Schulman_IJRR2014,
Zucker_IJRR2013,Betts_SIAM2010} can be applied to remarkably-complex
problems, although they may not converge to feasible solutions. A
widely used alternative is to rely on sampling-based
approaches~\cite{Choset_05,Lavalle_06}. These methods can cope with
high-dimensional problems, and guarantee to find a feasible solution,
if it exists and enough computing time is available. The RRT
method~\cite{LaValle_IJRR01} stands out among them, due to its
effectiveness and conceptual simplicity. However, it is well known
that RRT planners can be inefficient in certain
scenarios~\cite{Cheng_PhD2005}. Part of the complexity arises from
planning in the state space instead of in the lower-dimensional
configuration space~\cite{Pham_IJRR2017}. Nevertheless, the main issue
of RRT approaches is the disagreement of the metric used to measure
the distance between two given states, and the actual cost of moving
between such states, which must comply with the vector fields defined
by the dynamic constraints of the system. Several extensions to the
basic RRT planner have been proposed to alleviate this
issue~\cite{Sucan_TRO2012,Ladd_WAFR2005,
Plaku_RSS07,Shkolnik_ICRA09,Kalisiak_PhD2008,
Cheng_IROS2001,Cheng_ICRA2002,Jaillet_IROS2011}. None of these
extensions, however, can deal with the implicitly-defined
configuration spaces that arise when the problem includes kinematic
constraints~\cite{Stilman_IROS07,Berenson_IJRR11,
Porta_IJRR2012,Jaillet_TRO2013}. When considering both kinematic and
dynamic constraints, the planning problem requires the solution of
differential algebraic equations (DAE). The algebraic equations derive
from the kinematic constraints and the differential ones reflect the
system dynamics.

From constrained multibody dynamics it is well known that, when
simulating a system's motion, it is advantageous to directly deal with
the DAE of the system, rather than converting it into its ODE
form~\cite{Petzold_PDNP1992}. Several techniques have been used to
this end~\cite{Bauchau_JCND2008,Laulusa_JCND2008}. In the popular
Baumgarte method the drift is alleviated with control
techniques~\cite{Baumgarte_CMAME1972}, but the control parameters are
problem dependent and there is no general method to tune them. Another
way to reduce the drift is to use violation suppression
techniques~\cite{Braun_CMAME2009,Blajer_MSD2002}, but they do not
guarantee a drift-free integration. A better alternative are the
methods relying on local parameterizations~\cite{Potra_JSM1991}, since
they cancel the drift to machine accuracy. To the best of our
knowledge, this approach has never been applied in the context of
kinodynamic planning. However, it nicely complements an existing
planning method for implicitly-defined configuration
spaces~\cite{Porta_IJRR2012,Jaillet_TRO2013}, which also relies on
local parameterizations. The planner introduced in this paper can be
seen as an extension of the latter method to also deal with dynamics,
or an extension of \cite{LaValle_IJRR01} to include kinematic
constraints.

\section{Problem formalization}  \label{sec:formalization}

A robot configuration is described by means of a tuple $\vr{q}$ 
of~$n_q$ generalized coordinates $q_1,\ldots,q_{n_q}$, which determine the
positions and orientations of all links at a given instant of time.
There is total freedom in choosing the form and dimension of~$\vr{q}$,
but it must describe one, and only one, configuration. In this paper
we restrict our attention to constrained robots, i.e., those in which
$\vr{q}$ must satisfy a system of $n_e$ nonlinear equations
\begin{equation}
\vr{\Phi}(\vr{q}) = \vr{0},
\label{eq:phi}
\end{equation}
which express all joint assembly, geometric, or contact constraints to
be taken into account, either inherent to the robot design or
necessary for task execution. The configuration space~$\C$ of the
robot, or C-space for short, is the nonlinear variety
\begin{equation*}
\C = \{ \vr{q} : \vr{\Phi}(\vr{q})=\vr{0} \},
\end{equation*}
which may be quite complex in general. Under mild conditions, however,
we can assume that the Jacobian $\mt{\Phi}_{\vr{q}}(\vr{q})$ is full
rank for all $\vr{q} \in \C$, so that $\C$ is a smooth manifold of
dimension $d_{\C} = n_q - n_e$. This assumption is common because
C-space singularities can be avoided by judicious mechanical
design~\cite{Bohigas_SPRINGER2016}, or through the addition of
singularity-avoidance constraints into
Eq.~\eqref{eq:phi}~\cite{Bohigas_TRO2013}.

By differentiating Eq.~\eqref{eq:phi} with respect to time we obtain
\begin{equation}
\vr{\Phi}_{\vr{q}}(\vr{q}) \; \dot{\vr{q}}=\vr{0},
\label{eq:dotphi}
\end{equation}
which provides, for a given $\vr{q} \in \C$, the
feasible velocity vectors of the robot. 

Let $\vr{F}(\vr{x})=\vr{0}$ denote the system formed by
Eqs.~\eqref{eq:phi} and~\eqref{eq:dotphi}, where
$\vr{x}=(\vr{q},\dot{\vr{q}})\in \R{2 n_{q}}$. Our planning problem will take place
in the state space
\begin{equation} 
\X = \{ \vr{x} : \vr{F}(\vr{x})=\vr{0} \},
\label{eq:manifold}
\end{equation}
which encompasses all possible mechanical states of the
robot~\cite{Lavalle_06}. The fact that $\mt{\Phi}_{\vr{q}}(\vr{q})$ is
full rank guarantees that $\X$ is also a smooth manifold, but now of
dimension $d_{\X} = 2 \; d_{\C}$. This implies that the tangent space
of $\X$ at $\vr{x}$,
\begin{equation}
	\TX{\vr{x}} = \{ \dot{\vr{x}} \in \R{2 n_{q}} : 
	\vr{F}_{\vr{x}} \; \dot{\vr{x}} = \vr{0} \}
	\label{eq:tangent_space}
\end{equation}
is well-defined and $d_{\X}$-dimensional for any $\vr{x} \in \X$.

We shall encode the forces and torques of the actuators into an action
vector $\vr{u}$ of dimension $n_u$. Our main interest will be on
fully-actuated robots, i.e., those for which $n_u = d_\C$, but the
developments that follow are also applicable to over- or
under-actuated robots.

Given a starting state $\vr{x}_s \in \X$, and the action vector as a
function of time,  $\vr{u} = \vr{u}(t)$, it is well-known that the
time evolution of the robot is determined by a DAE of the form
\begin{eqnarray}
\vr{F}(\vr{x})=\vr{0} \label{eq:DAE1} \\ 
\dot{\vr{x}} = \vr{g}(\vr{x},\vr{u}) \label{eq:DAE2}
\end{eqnarray}
The first equation forces the states $\vr{x}$ to lie in $\X$. The
second equation models the dynamics of the system~\cite{Lavalle_06},
which can be formulated, e.g., using the multiplier form of the
Euler-Lagrange equations~\cite{Potra_JSM1991}. 
For each value of $\vr{u}$, it defines a vector field over $\X$, which
can be used to integrate the robot motion forward in time, using
proper numerical methods.

Since in practice the actuator forces are limited, $\vr{u}$ is always
constrained to take values in some bounded subset $\U$ of $\R{n_u}$,
which limits the range of possible state velocities $\dot{\vr{x}} =
\vr{g}(\vr{x},\vr{u})$ at each $\vr{x} \in \X$. During its motion,
moreover, the robot cannot incur in collisions with itself or with the
environment, constraining the feasible states $\vr{x}$ to those lying
in a subset $\Xfree \subseteq \X$ of non-collision states.

With the previous definitions, the planning problem we confront can be
phrased as follows. Given two states of $\Xfree$,~$\vr{x}_s$ and
$\vr{x}_g$, find an action trajectory $\vr{u} = \vr{u}(t) \in \U$ such
that the trajectory $\vr{x} = \vr{x}(t)$ with $\vr{x}(0) = \vr{x}_s$
of the system determined by~Eqs.~\eqref{eq:DAE1} and~\eqref{eq:DAE2},
fulfills $\vr{x}(t_f) = \vr{x}_g$ for some time $t_f > 0$, and
$\vr{x}(t) \in \Xfree$ for all $t \in [0,t_f]$.

\section{Mapping and Exploring the State Space}
\label{sec:state_space_atlas}

The fact that $\X$ is an implicitly defined manifold complicates the
design of an RRT planner able to solve the previous problem. In
general, $\X$ does not admit a global parameterization and there is no
straightforward way to sample~$\X$ uniformly. The integration of
Eq.~\eqref{eq:DAE2}, moreover, will yield robot trajectories drifting
away from $\X$ if numerical methods for plain ODE systems are used.
Even so, we next see that both issues can be circumvented by using an
atlas of~$\X$. If built up incrementally, such an atlas will lead to
an efficient means of extending an RRT over the state space.

\subsection{Atlas construction}
\label{subsec:atlas_construction}

\begin{figure}[t!]	
	\begin{center}
		\pstool[width=0.9\linewidth]{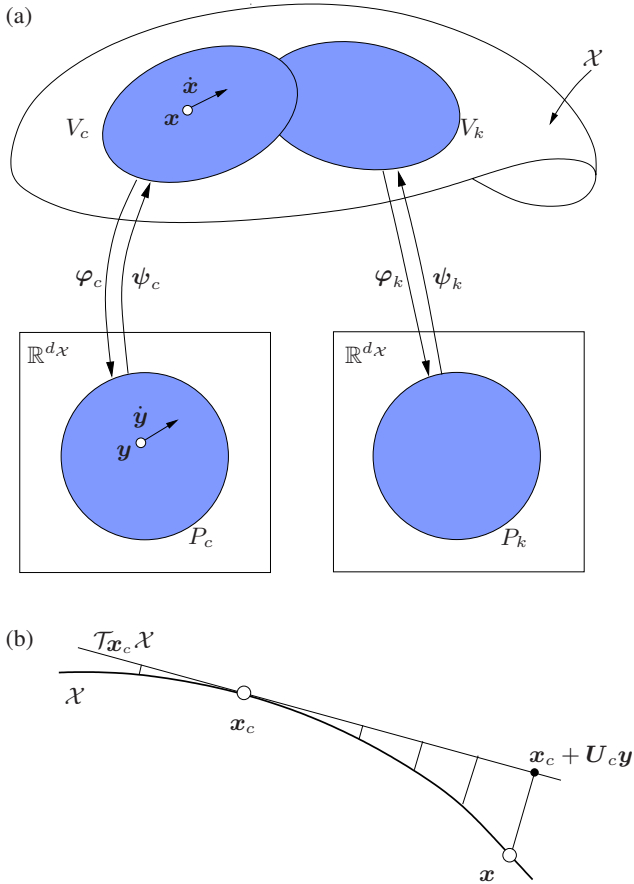}{
			\psfrag{X}[l]{\small $\X$}
			\psfrag{(a)}[l]{\small (a)}
			\psfrag{(b)}[l]{\small (b)}
			\psfrag{x}[l]{\small $\vr{x}$}
			\psfrag{y}[l]{\small $\vr{y}$}
			\psfrag{xdot}[l]{\small $\dot{\vr{x}}$}
			\psfrag{ydot}[l]{\small $\dot{\vr{y}}$}
			\psfrag{R}[l]{\small $\R{d_{\X}}$}
			\psfrag{Ui}[l]{\small $P_c$}
			\psfrag{Vi}[l]{\small $V_c$}
			\psfrag{Uj}[l]{\small $P_k$}
			\psfrag{Vj}[l]{\small $V_k$}
			\psfrag{Vj}[l]{\small $V_k$}
			\psfrag{psii}[l]{\small $\vr{\psi}_c$}
			\psfrag{psij}[l]{\small $\vr{\psi}_k$}
			\psfrag{phii}[l]{\small $\vr{\varphi}_c$}
			\psfrag{phij}[l]{\small $\vr{\varphi}_k$}
			\psfrag{xz}[l]{\small $\vr{x}$}	
			\psfrag{T}[l]{\small $\TX{\vr{x}_c}$}	 		
			\psfrag{xt}[l]{\small $\vr{x}_c+\mt{U}_c\vr{y}$}	
			\psfrag{x0}[l]{\small $\vr{x}_c$}	
		}
	\end{center}
	\vspace{-0.25cm} \caption{\label{fig:charts}(a) An atlas is a
		collection of maps $\vr{\varphi}$ providing local coordinates to
		all points of $\X$. The inverse maps $\vr{\psi}$ convert the vector
		fields on $\X$ to vector fields on $\R{d_{\X}}$. (b) The projection
		of the points $\vr{x}\in\X$ to $\TX{\vr{x}_c}$ leads to specific
		instances of $\vr{\varphi}_c$ and $\vr{\psi}_c$.}
\end{figure}

Formally, an atlas of $\X$ is a collection of charts mapping~$\X$
entirely, where each chart $c$ is a local diffeomorphism
$\vr{\varphi}_c$ from an open set $V_c \subset \X$ to an open set $P_c
\subseteq \R{d_{\X}}$ [Fig.~\ref{fig:charts}(a)]. The~$V_c$ sets can
be thought of as partially-overlapping tiles covering~$\X$, in such a
way that every $\vr{x} \in \X$ lies in at least one~$V_c$. The point
$\vr{y}=\vr{\varphi}_c(\vr{x})$ provides the local coordinates, or
parameters, of $\vr{x}$ in chart $c$. Since each $\vr{\varphi}_c$ is a
diffeomorphism, its inverse map ${\vr{\psi}_c}=\vr{\varphi}_c\inv$
exists and  gives a local parameterization of $V_c$.

To construct $\vr{\varphi}_c$ and $\vr{\psi}_c$ we shall use the
so-called tangent space parameterization~\cite{Potra_JSM1991}. In this
approach, the map \mbox{$\vr{y} = \vr{\varphi}_c(\vr{x})$} around a given
\mbox{$\vr{x}_c \in \X$} is obtained by projecting~$\vr{x}$
orthogonally to $\TX{\vr{x}_c}$ [Fig.~\ref{fig:charts}(b)]. Thus
$\vr{\varphi}_c$ becomes
\begin{equation}\label{eq:phic}
\vr{y} = \mt{U}_c\tr \; (\vr{x}-\vr{x}_c),
\end{equation}
where $\mt{U}_c$ is a $2 n_q \times d_\X$ matrix whose columns provide
an orthonormal basis of $\TX{\vr{x}_c}$. $\mt{U}_c$ can be computed efficiently using
the QR decomposition of $\mt{F}_{\vr{x}_c}$. The inverse map \mbox{$\vr{x}=\vr{\psi}_c(\vr{y})$} is implicitly determined by the system of
nonlinear equations
\begin{equation}
\begin{array}{c}
\vr{F}(\vr{x})=\vr{0}, \\
\mt{U}_c\tr(\vr{x}-\vr{x}_c)-\vr{y} = \vr{0}.
\end{array}
\label{eq:TS_param}
\end{equation}
For a given $\vr{y}$, these equations can be solved for $\vr{x}$ by
means of the Newton-Raphson method.

Assuming that an atlas has been created, the problem of sampling $\X$
boils down to sampling the~$P_c$ sets, since the $\vr{y}$ values can
always be projected to $\X$ using the corresponding map
\mbox{$\vr{x}=\vr{\psi}_c(\vr{y})$}. Also, the atlas allows the
conversion of the vector field defined by Eq.~\eqref{eq:DAE2} into one
in the coordinate spaces~$P_c$. The time derivative of
Eq.~\eqref{eq:phic}, \mbox{$\dot{\vr{y}}=\mt{U}_c\tr \dot{\vr{x}}$},
gives the relationship between the two vector fields, and allows
writing
\begin{equation}
\dot{\vr{y}}= \mt{U}_c\tr \: \vr{g}(\vr{\psi}_c(\vr{y}),\vr{u}), 
\label{eq:ODE_param}
\end{equation}
which is Eq.~\eqref{eq:DAE2} but expressed in local coordinates. This
equation forms the basis of the so-called tangent-space
parameterization methods for the integration of DAE
systems~\cite{Hairer_NM2001,Hairer_SPRINGER2006}. Given a state
$\vr{x}_k$ and an action $\vr{u}$, $\vr{x}_{k+1}$ is estimated by
obtaining $\vr{y}_k = \vr{\varphi}_c(\vr{x}_k)$, then computing
$\vr{y}_{k+1}$ using a discrete form of Eq.~\eqref{eq:ODE_param}, and
finally getting \mbox{$\vr{x}_{k+1} = \vr{\psi}_c({\vr{y}_{k+1}})$}.
The procedure guarantees that~$\vr{x}_{k+1}$ will lie on $\X$, which
makes the integration compliant with all kinematic constraints in
Eq.~\eqref{eq:DAE1}.

\subsection{Incremental atlas and RRT expansion}
\label{subsec:increm_construction}

One could build a full atlas of the implicitly-defined state
space~\cite{Henderson_BC02} and then use its local parameterizations
to define a kinodynamic RRT. However, the construction of a complete
atlas is only feasible for low-dimensional state spaces. Moreover,
only part of the atlas is necessary to solve a given motion planning
problem. Thus, a better alternative is to combine the construction of
the atlas and the expansion of the RRT~\cite{Jaillet_TRO2013}. In this
approach, a partial atlas is used to generate random states and to add
branches to the RRT. Also, as described next, new charts are created
as the RRT branches reach unexplored areas of the state space.

Suppose that $\vr{x}_k$ and $\vr{x}_{k+1}$ are two consecutive steps
along an RRT branch whose parameters in the chart defined
at~$\vr{x}_c$ are $\vr{y}_k$ and $\vr{y}_{k+1}$, respectively. Then, a
new chart at $\vr{x}_k$  is generated if any of the following
conditions holds
\begin{align}
  \|\vr{x}_{k+1}- (\vr{x}_c +\mt{U}_c\:\vr{y}_{k+1}) \| &> \epsilon, \\
  \frac{\| \vr{y}_{k+1}-\vr{y}_k\|}{\| \vr{x}_{k+1}-\vr{x}_k\|} &< cos(\alpha), \\
  \|\vr{y}_{k+1}\| &> \rho, \label{eq:rho}
\end{align}
where $\epsilon$, $\alpha$, and $\rho$ are user-defined parameters.
The three conditions are introduced to ensure that the chart
domains~$P_c$ capture the overall shape of~$\X$ with sufficient
detail. The first condition limits the maximal distance between the
tangent space and the manifold. The second condition ensures a bounded
curvature in the part of the manifold covered by a local
parameterization, as well as a smooth transition between charts.
Finally, the third condition is introduced to ensure the generation of
new charts as the RRT grows, even for (almost) flat manifolds.
\begin{figure}[t!]	
	\begin{center}
		\pstool[width=0.9\linewidth]{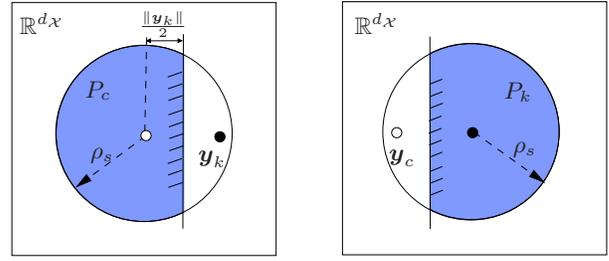}{
			\psfrag{norm}[l]{\tiny $\frac{\Vert \vr{y}_k\Vert}{2}$}
			\psfrag{r}[l]{\small $\rho_s$}
			\psfrag{yk}[l]{\small $\vr{y}_k$}
			\psfrag{yc}[l]{\small $\vr{y}_c$}
			\psfrag{R}[l]{\small $\R{d_{\X}}$}
			\psfrag{Ui}[l]{\small $P_c$}
			\psfrag{Vi}[l]{\small $V_c$}
			\psfrag{Uj}[l]{\small $P_k$}
		}
	\end{center}
	\vspace{-0.25cm} \caption{\label{fig:bounding}Bounding of the
	parameter sets $P_c$ and $P_k$ of the two neighboring charts in
	Fig.~\ref{fig:charts}. Note that $\vr{y}_c=\vr{\varphi}_k(\vr{x}_c)$ and
	$\vr{y}_k=\vr{\varphi}_c(\vr{x}_k)$.}
\end{figure}

\begin{algorithm}[t]
  \begin{small}
    \BlankLine
    \textsc{\bf Planner}($\vr{x}_s,\vr{x}_g$)\\
    \LinesNumbered
    \DontPrintSemicolon
    \SetKwInOut{Input}{input} \SetKwInOut{Output}{output}
    \Input{The query states, $\vr{x}_s$ and $\vr{x}_g$.}
    \Output{A trajectory connecting $\vr{x}_s$ and $\vr{x}_g$.}
    $T_s \leftarrow \textsc{InitRRT}(\vr{x}_s)$ \nllabel{alg:RRTs} \\
    $T_g \leftarrow \textsc{InitRRT}(\vr{x}_g)$ \nllabel{alg:RRTg} \\
    $A \leftarrow \textsc{InitAtlas}(\vr{x}_s,\vr{x}_g)$ \nllabel{alg:initAtlas}\\
    \Repeat{$\|\vr{x}_l-\vr{x}_l'\|<\beta$ \nllabel{alg:mainLoopStart}} 
    {
      $\vr{x}_r \leftarrow \textsc{Sample}(A,T_s)$ \nllabel{alg:sample}\\
      $\vr{x}_n \leftarrow \textsc{NearestState}(T_s,\vr{x}_r)$ \nllabel{alg:nn1}\\
      $\vr{x}_l \leftarrow \textsc{ExtendRRT}(A,T_s,\vr{x}_n,\vr{x}_r)$ \nllabel{alg:extend1}\\
      $\vr{x}_n' \leftarrow \textsc{NearestState}(T_g,\vr{x}_l)$ \nllabel{alg:nn2}\\
      $\vr{x}_l' \leftarrow \textsc{ExtendRRT}(A,T_g,\vr{x}_n',\vr{x}_l)$ \nllabel{alg:extend2}\\
      \textsc{Swap}$(T_s,T_g)$ \nllabel{alg:swap} \nllabel{alg:mainLoopEnd}\\
    }
    \textsc{Return}(\textsc{Trajectory}($T_s,\vr{x}_l,T_g,\vr{x}_l'$)) \nllabel{alg:path}
  \end{small}
  \caption{The main procedure of the planner} \label{alg:main}
\end{algorithm}

\subsection{Chart coordination}
\label{subsec:chart_cooord}

Since the charts will be used to sample the state space uniformly, it
is important to reduce the overlap between new charts and those
already in the atlas. Otherwise, the areas of $\X$ covered by several
charts would be oversampled. To this end, the set of valid parameters
for each chart~$c$,~$P_c$, is represented as the intersection of a
ball of radius $\rho_s$ and a number of half-planes, all defined in
$\TX{\vr{x}_c}$. The set $P_c$ is progressively bounded as new
neighboring charts are created around chart $c$. If, while growing an
RRT branch using the local parameterization provided by
$\TX{\vr{x}_c}$, a chart is created  on a point $\vr{x}_k$ with
parameter vector~$\vr{y}_k$ in $P_c$, then  the following inequality
\begin{equation}
  \vr{y}\tr \vr{y}_k  -  \frac{\|\vr{y}_k\|^2}{2}  \leq 0 \label{eq:setF}
\end{equation}
with $\vr{y} \in \R{d_\X}$, is added to the definition of $P_c$
(Fig.~\ref{fig:bounding}). A similar inequality is added to $P_k$, the
chart at $\vr{x}_k$, by projecting~$\vr{x}_c$ to $\TX{\vr{x}_k}$. The
parameter $\rho_s$ must be larger than~$\rho$ to guarantee that the
RRT branches in chart $c$ will eventually trigger the generation of
new charts, i.e., to guarantee that Eq.~\eqref{eq:rho} eventually
holds.

\begin{algorithm}[t]
  \begin{small}
    \BlankLine
    \textsc{\bf ExtendRRT}($A,T,\vr{x}_n,\vr{x}_r$)\\
    \LinesNumbered
    \SetKwInOut{Input}{input} \SetKwInOut{Output}{output}
    \Input{An atlas, $A$, a tree, $T$, the state from
      where to extend the tree, $\vr{x}_n$, and the
      random sample to be reached, $\vr{x}_r$.}
    \Output{The updated tree.}
    \BlankLine
    $d_b \leftarrow \infty$  \\
    \ForEach{$\vr{u} \in \mathcal{U}$ \label{alg:allActions}}
    {
      $\vr{x} \leftarrow \textsc{SimulateAction}(A,T,\vr{x}_n,\vr{x}_r,\vr{u})$ \label{alg:simulateAction} \\
      $d \leftarrow \|\vr{x}-\vr{x}_r\|$ \\
      \If{$d<d_b$}
      {
        $\vr{x}_b \leftarrow \vr{x}$\\
        $\vr{u}_b \leftarrow \vr{u}$\\
        $d_b \leftarrow d$\\
      }
    } \label{alg:simulateAction2} 
    \If{$\vr{x}_b \notin T$}
    {
      $T \leftarrow \textsc{AddActionState}(T,\vr{x}_n,\vr{u}_b,\vr{x}_b)$ \label{alg:add2Tree}
    }    
  \end{small}
  \caption{Extend an RRT.}
  \label{alg:extend}
\end{algorithm}

\section{The Planner} 
\label{sec:theplanner}

\subsection{Higher-level structure} 
\label{subsec:higher-level}

Algorithm~\ref{alg:main} gives the high level pseudocode of the
planner. It implements a bidirectional RRT where one tree is extended
(line~\ref{alg:extend1}) towards a random sample (generated in
line~\ref{alg:sample}) and then the other tree is extended
(line~\ref{alg:extend2}) towards the state just added to the first
tree. The process is repeated until the trees become connected with a
given user-specified accuracy (parameter $\beta$ in
line~\ref{alg:mainLoopStart}). Otherwise, the trees are swapped
(line~\ref{alg:swap}) and the process is repeated. Tree extensions are
always initiated at the state in the tree closer to the target state
(lines~\ref{alg:nn1} and~\ref{alg:nn2}). Different metrics can be used
without affecting the overall structure of the planner. For
simplicity, the Euclidean distance in state space  is used in the
approach presented here. The main difference of this algorithm with
respect to the standard bidirectional RRT is that here we use an atlas
(initialized in line~\ref{alg:initAtlas}) to parameterize the state
space manifold.

Algorithm~\ref{alg:extend} provides the pseudocode of the procedure to
extend an RRT from a given state $\vr{x}_n$ towards a goal
state~$\vr{x}_r$. The procedure simulates the motion of the system
(line~\ref{alg:simulateAction}) for a set of actions, which can be
selected at random or taken from a predefined set
(line~\ref{alg:allActions}). The action that yields a new state closer
to $\vr{x}_r$ is added to the RRT  with an edge connecting it to
$\vr{x}_n$ (line~\ref{alg:add2Tree}). The action generating the
transition from~$\vr{x}_n$ to the new state is also stored in the tree
so that action trajectory can be returned after planning.

\begin{algorithm}[t]
  \begin{small}
    \BlankLine
    \textsc{\bf Sample}($A, T$)\\
    \LinesNumbered
    \DontPrintSemicolon
    \SetKwInOut{Input}{input} \SetKwInOut{Output}{output}
    \Input{The atlas, $A$, the tree currently extended, $T$.}
    \Output{A sample on the atlas.}
    \Repeat{$\vr{y}_r \in P_r$ \nllabel{alg:untilP}} 
    {
      $r \leftarrow \textsc{RandomChartIndex}(A, T)$ \nllabel{alg:randomChart}\\
      $\vr{y}_r \leftarrow \textsc{RandomOnBall}(\rho_s)$ \nllabel{alg:sampleInBall}\\
    }
    \textsc{Return}($\vr{x}_r + \mt{U}_r \:\vr{y}_r$) \nllabel{alg:returnRandom}
  \end{small}
  \caption{Sample a state.} \label{alg:sampling}
\end{algorithm}

\subsection{Sampling}
\label{subsec:sampling}

Algorithm~\ref{alg:sampling} describes the procedure to generate
random states. First, one of the charts covering the tree to be
expanded is selected at random (line~\ref{alg:randomChart}) and then a
vector of parameters is generated randomly in a ball of radius
$\rho_s$ (line~\ref{alg:sampleInBall}). The sampling process is
repeated until the parameters are inside the set $P_c$ for the
selected chart. Finally, the sampling procedure returns the ambient
space coordinates corresponding to the randomly generated parameters
(line~\ref{alg:returnRandom}).

\subsection{Dynamic simulation}
\label{subsec:simulation}

In order to simulate the system evolution from a given
state~$\vr{x}_k$, the DAE system is treated as an ODE on the
manifold~$\X$, as described in Section~\ref{sec:state_space_atlas}.
Any numerical integration method, either explicit or implicit, could
be used to obtain a solution to Eq.~\eqref{eq:ODE_param} in the
parameter space, and then solve Eq.~\eqref{eq:TS_param} to transform
back to the manifold. However, in this planner an implicit integrator,
the trapezoidal rule, is used, as its computational cost (integration
and projection to the manifold) is similar to the cost of using an
explicit method of the same order \cite{Potra_JSM1991}. Moreover, it
gives more stable and accurate solutions over long time intervals.
Using this rule, Eq.~\eqref{eq:ODE_param} is discretized as
\begin{equation} \label{eq:discrete_ODE}
\vr{y}_{k+1}=  \vr{y}_{k}+ \frac{h}{2}\mt{U}_c\tr \:(\vr{g}(\vr{x}_k,\vr{u})+\vr{g}(\vr{x}_{k+1},\vr{u})),
\end{equation}
where $h$ is the integration time step. Notice that this rule is
symmetric and, thus, it can be used to obtain time reversible
solutions~\cite{Hairer_NM2001,Hairer_SPRINGER2006}. This property is
specially useful in our planner, since the tree with root at
$\vr{x}_g$ is built backwards in time. The value~$\vr{x}_{k+1}$ in
Eq.~\eqref{eq:discrete_ODE} is still unknown, but it can be obtained
by using Eq.~\eqref{eq:TS_param} as
\begin{equation}
\begin{array}{c}
\vr{F}(\vr{x}_{k+1})=\vr{0}, \\
\mt{U}_c\tr(\vr{x}_{k+1}-\vr{x}_c)-\vr{y}_{k+1} = \vr{0}.
\end{array}
\label{eq:TS_param_discrete}
\end{equation}
Now, both Eq.~\eqref{eq:discrete_ODE} and
Eq.~\eqref{eq:TS_param_discrete} are combined to form the following
system of equations
\begin{equation}
\begin{small}
\begin{array}{c}
\vr{F}(\vr{x}_{k+1})=\vr{0}, \\
\hspace{-0.2cm}
\mt{U}_c\tr(\vr{x}_{k+1}- \frac{h}{2} (\vr{g}(\vr{x}_k,\vr{u})+\vr{g}(\vr{x}_{k+1},\vr{u}))-\vr{x}_c)- \vr{y}_k = \vr{0},
\end{array}
\label{eq:integration}
\end{small}
\end{equation}
where $\vr{x}_k$, $\vr{y}_k$, and $\vr{x}_c$ are known
and~$\vr{x}_{k+1}$ is the unknown to determine. Any Newton method can
be used to solve this system, but the Broyden method is particularly
adequate since it avoids the computation of the Jacobian of the system
at each step. \citet{Potra_JSM1991} gave an approximation of this
Jacobian, that allowed $\vr{x}_{k+1}$ to be found in few iterations.

\begin{algorithm}[t]
  \begin{small}
    \BlankLine
    \textsc{\bf SimulateAction}($A,T,\vr{x}_k,\vr{x}_g,\vr{u}$)\\
    \LinesNumbered
    \SetKwInOut{Input}{input} \SetKwInOut{Output}{output}
    \Input{An atlas, $A$, a tree, $T$, the 
           state from where to start the simulation, $\vr{x}_k$, the
           state to approach $\vr{x}_g$, and the action 
           to simulate, $\vr{u}$.}
    \Output{The last state in the simulation.}
    \BlankLine
    
    $c \leftarrow \textsc{ChartIndex}(\vr{x}_k)$ \nllabel{alg:chart} \\
    $\textsc{Feasible} \leftarrow \textsc{True}$\\
    $t \leftarrow 0$\\
    \While{\textsc{Feasible} {\bf and} $\|\vr{x}_k-\vr{x}_g\| > \delta$ {\bf and} $|t|\leq t_m$ \nllabel{alg:loopStart}}
    {
      $\vr{y}_{k} \leftarrow \vr{\varphi}_c(\vr{x}_{k})$ \nllabel{alg:un}\\
      $(\vr{x}_{k+1},\vr{y}_{k+1},h) \leftarrow \textsc{NextState}(\vr{x}_{k},\vr{y}_{k},\vr{u},\vr{F},\mt{U}_c,\delta)$ \label{alg:nextStep}\\
      \If{\textsc{Collision}$(\vr{x}_{k+1})$ {\bf or} \textsc{OutOfWorkspace}$(\vr{x}_{k+1})$ \label{alg:obstacle}}
      {
        $\textsc{Feasible} \leftarrow \textsc{False}$ \nllabel{alg:collision}
      }
      \Else 
      {
        \If{$\| \vr{x}_{k+1} - (\vr{x}_c +\mt{U}_c \:\vr{y}_{k+1}) \| > \epsilon$ {\bf or} 
            $\| \vr{y}_{k+1} -\vr{y}_{k} \| / \| \vr{x}_{k+1}-\vr{x}_{k} \| < \cos(\alpha)$ {\bf or}
            $\| \vr{y}_{k+1} \| >  \rho$ \nllabel{alg:newChart}}
        {
          $c \leftarrow \textsc{AddChartToAtlas}(A,\vr{x}_k)$ \nllabel{alg:createChart}\\
        }
        \Else
        {
          \If{$\vr{y}_{k+1} \notin P_c$ \nllabel{alg:neighbourChart}}
          {
            $c \leftarrow \textsc{NeighborChart}(A,c,\vr{y}_{k+1})$ \nllabel{alg:moveChart}\\
          }
          $t \leftarrow t + h$\\
          $\vr{x}_k \leftarrow \vr{x}_{k+1}$\\
        }
      }
    }
    \textsc{Return}($\vr{x}_k$) \nllabel{alg:returnLast}
  \end{small}
  \caption{Simulate an action.}
 \label{alg:simulate}
\end{algorithm}
Algorithm~\ref{alg:simulate} summarizes the procedure to simulate a
given action, $\vr{u}$ from a particular state, $\vr{x}_k$. The
simulation is carried on while the path is not blocked by an obstacle
or by a workspace limit (line~\ref{alg:obstacle}), while the goal
state is not reached (with accuracy~$\delta$), or for a maximum time
span,~$t_m$ (line~\ref{alg:loopStart}). At each simulation step, the
key procedure is the solution of Eq.~\eqref{eq:integration}
(line~\ref{alg:nextStep}), which provides the next state,
$\vr{x}_{k+1}$, given the current one, $\vr{x}_k$, the corresponding
vector of parameters~$\vr{y}_k$, the action to simulate,~$\vr{u}$, the
orthonormal basis of $\TX{\vr{x}_c}$ for the chart including
$\vr{x}_k$,~$\mt{U}_c$, and the desired step size in tangent space,
$\delta$. For backward integration, i.e., when extending the RRT with
root at $\vr{x}_g$, the time step~$h$ in Eq.~\eqref{eq:integration} is
negative. In any case, $h$ is adjusted so that the change in parameter
space, $\|\vr{y}_{k+1}-\vr{y}_{k}\|$, is bounded by $\delta$, with
$\delta \ll \rho$. This is necessary to detect the transitions between
charts which can occur either because the next state triggers the
creation of a new chart (line~\ref{alg:createChart}) or because it is
not in the part of the manifold covered by the current chart
(line~\ref{alg:neighbourChart}) and, thus, it is in the part covered
by a neighboring chart (line~\ref{alg:moveChart}).
 
\begin{figure}[t]
	\begin{center}
		\includegraphics[width=0.4\linewidth]{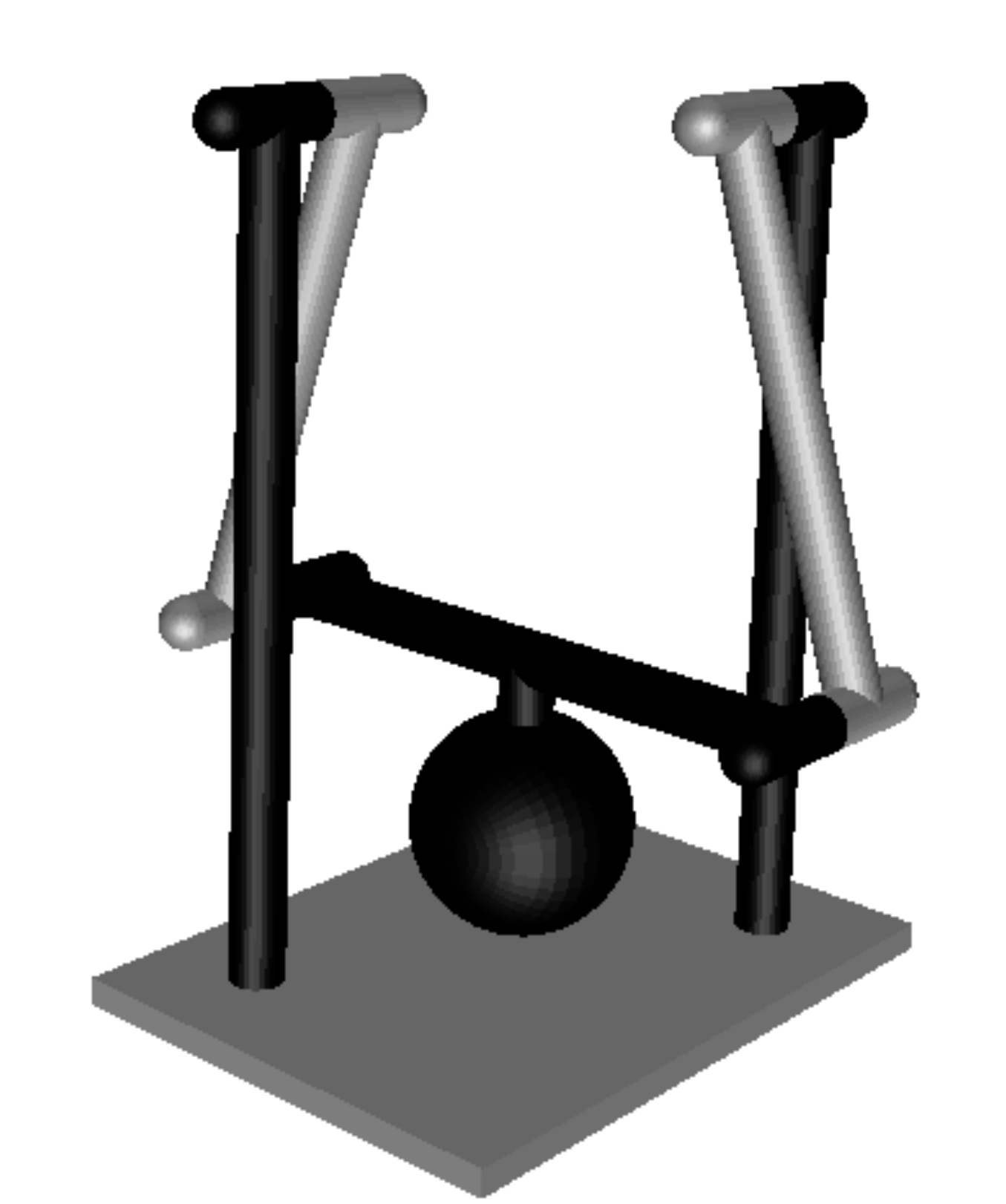}%
		\includegraphics[width=0.5\linewidth]{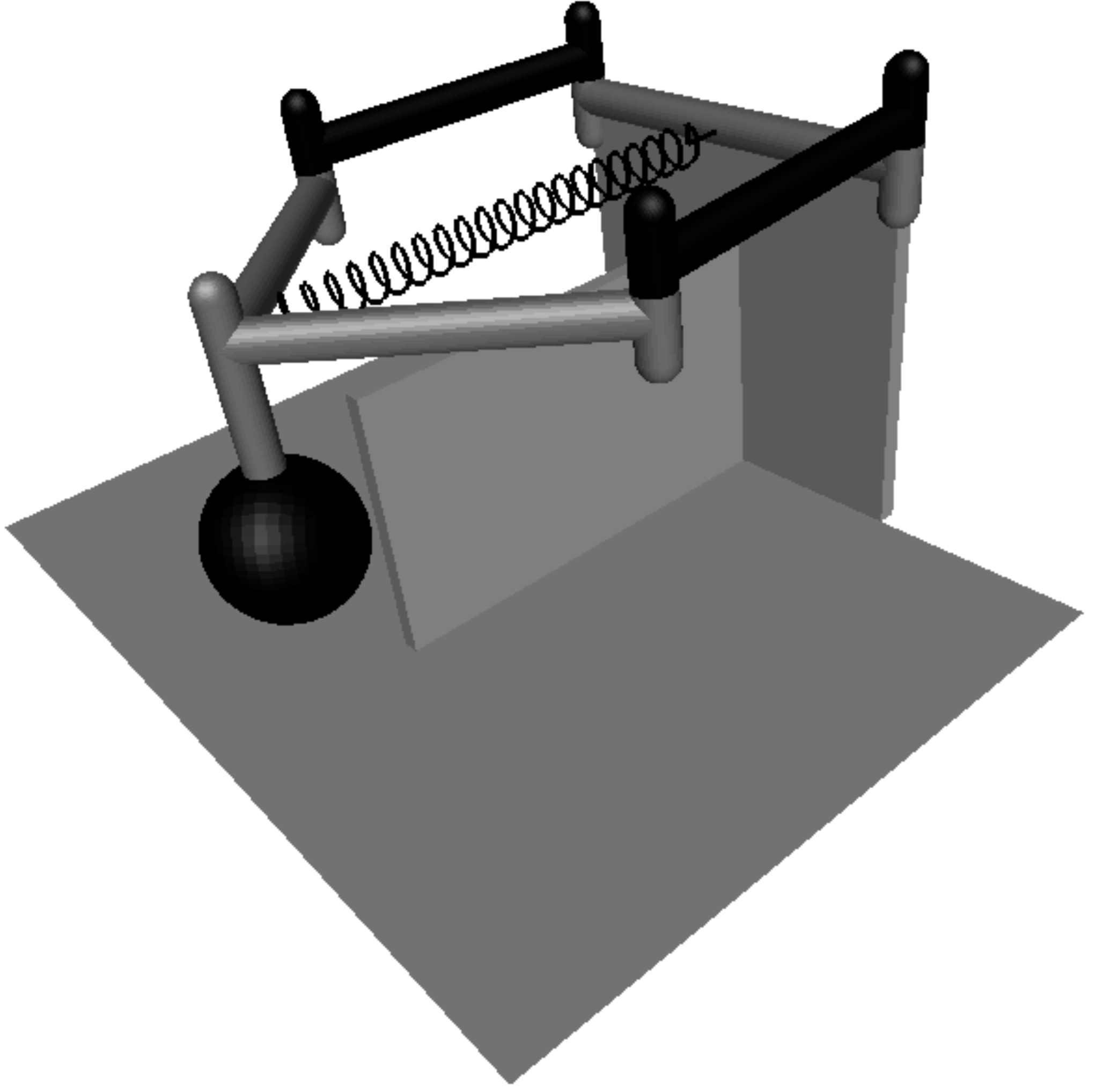}
		\includegraphics[width=0.9\linewidth]{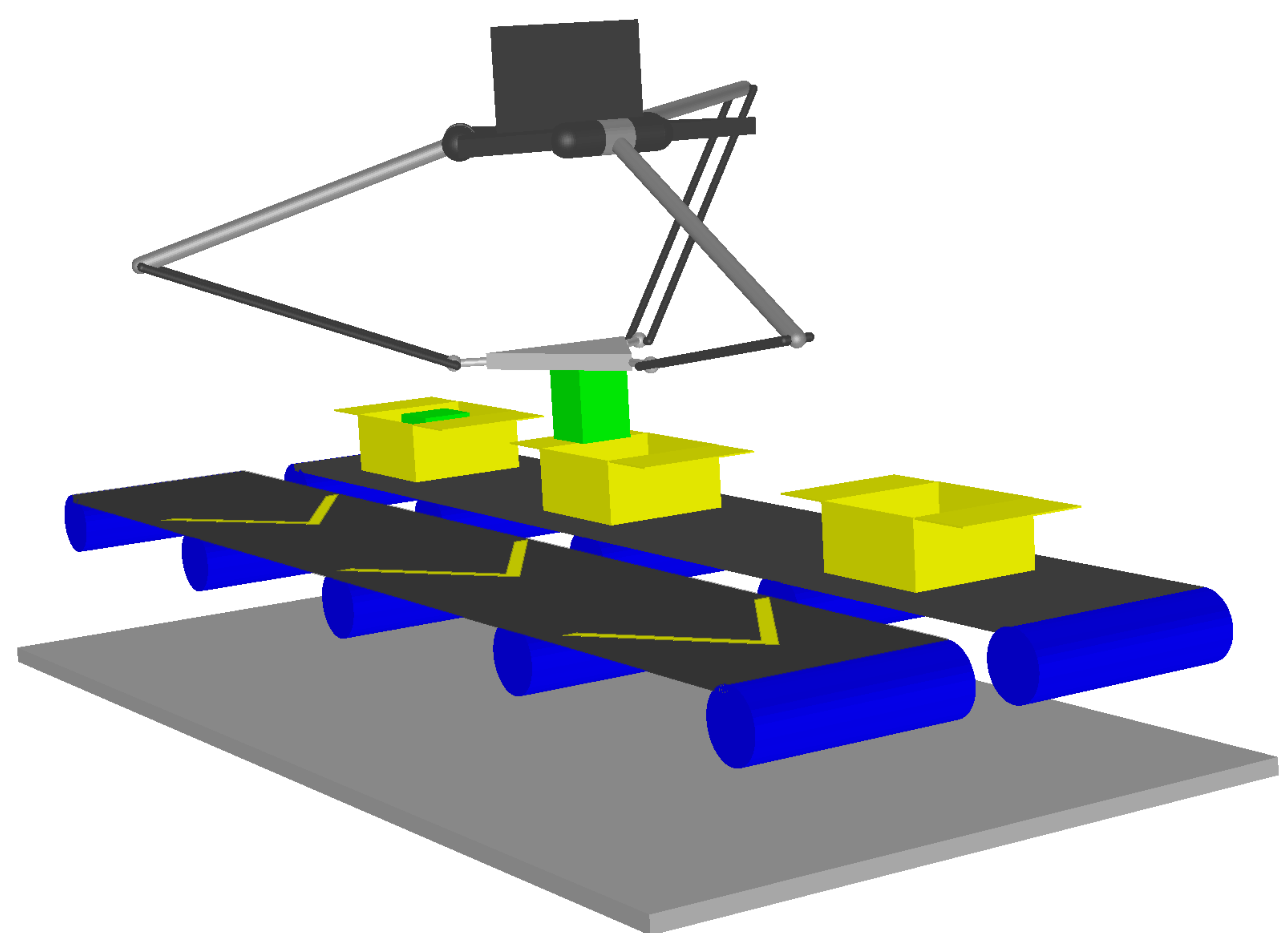}\\[0.5cm]
	\end{center}
	\caption{Test cases used to validate the planner: a four-bar pendulum (top-left), a five-bar robot (top-right), and a Delta robot (bottom).} \label{fig:examples}
\end{figure}

\section{Test Cases} 
\label{sec:experiments}

The planner has been implemented in C and integrated in the CUIK Suite \cite{Porta_RAM2014}. We next illustrate its
performance in three test cases of increasing complexity
(Fig.~\ref{fig:examples}). The first case was already mentioned in the
introduction. It consists of a planar four-bar pendulum with limited
motor torque that has to move a load. The robot may need to oscillate
several times to move between the two states shown in
Fig.~\ref{fig:pendulum_mech}. The second test case involves a planar
five-bar robot equivalent to the Dextar prototype
\cite{Bourbonnais_TMEC2015}, but with an added spring to enhance its
compliance. The goal here is to move the load from one side to the
other of a wall, with null initial and final velocities. Unlike in the
first case, collisions may occur here, and thus they should be
avoided. In the third case, a Delta robot moves a heavy load in a
pick-and-place scenario. It picks up the load from a conveyor belt
moving at constant velocity, and places it at rest inside a box on a
second belt. In contrast to typical Delta robot applications, here the
weight of the load is considerable, which increases dynamics effects
substantially. Table~\ref{tab:param} summarizes the problem
dimensions, parameters, and performance statistics of all test cases.
The full set of geometric and dynamic parameters will be made
available online in the final version of the paper.

\begin{table*}[t]
	\caption{Test case dimensions, parameters, and performance statistics. \label{tab:param}} 
	\begin{center}
		\begin{tabular}{ccccccccccc}
			\toprule
			\textbf{Robot} & $n_q$ & $n_{e}$ & $d_{\C}$ & $d_{\X}$ &No. of actions & $\tau_{max}$ [Nm] &  $\beta$ & No. of samples & No. of charts & Exec. Time [s] \\ 
			\midrule
			\multicolumn{1}{c}{\multirow{4}{*}{Four-Bars}} & 4 & 3&1   & 2 & 3 & 16 & 0.1  & 452 & 122 & 2.7\\
			& 4 & 3 & 1 & 2 & 3& 12&0.1  & 569 & 145 & 3.4 \\
			& 4 & 3 & 1 & 2 & 3 & 8 &0.1 & 1063 & 195 & 6.3 \\
			& 4 & 3 & 1 & 2& 3& 4 &0.1 & 2383 & 248 & 14.2  \\
			\midrule		
			Five-Bars & 5 & 3 & 2 & 4 &5& 0.1  &0.25  &  15980 & 101 & 124 \\ 
			\midrule	
			Delta  & 15 & 12 & 3 & 6 & 7& 1 &0.5 & 1654  & 58 & 183 \\ 
			\bottomrule
		\end{tabular}
	\end{center}
\end{table*}
\begin{figure*}[t!]
	\begin{center}
		\includegraphics[width=0.24\linewidth]{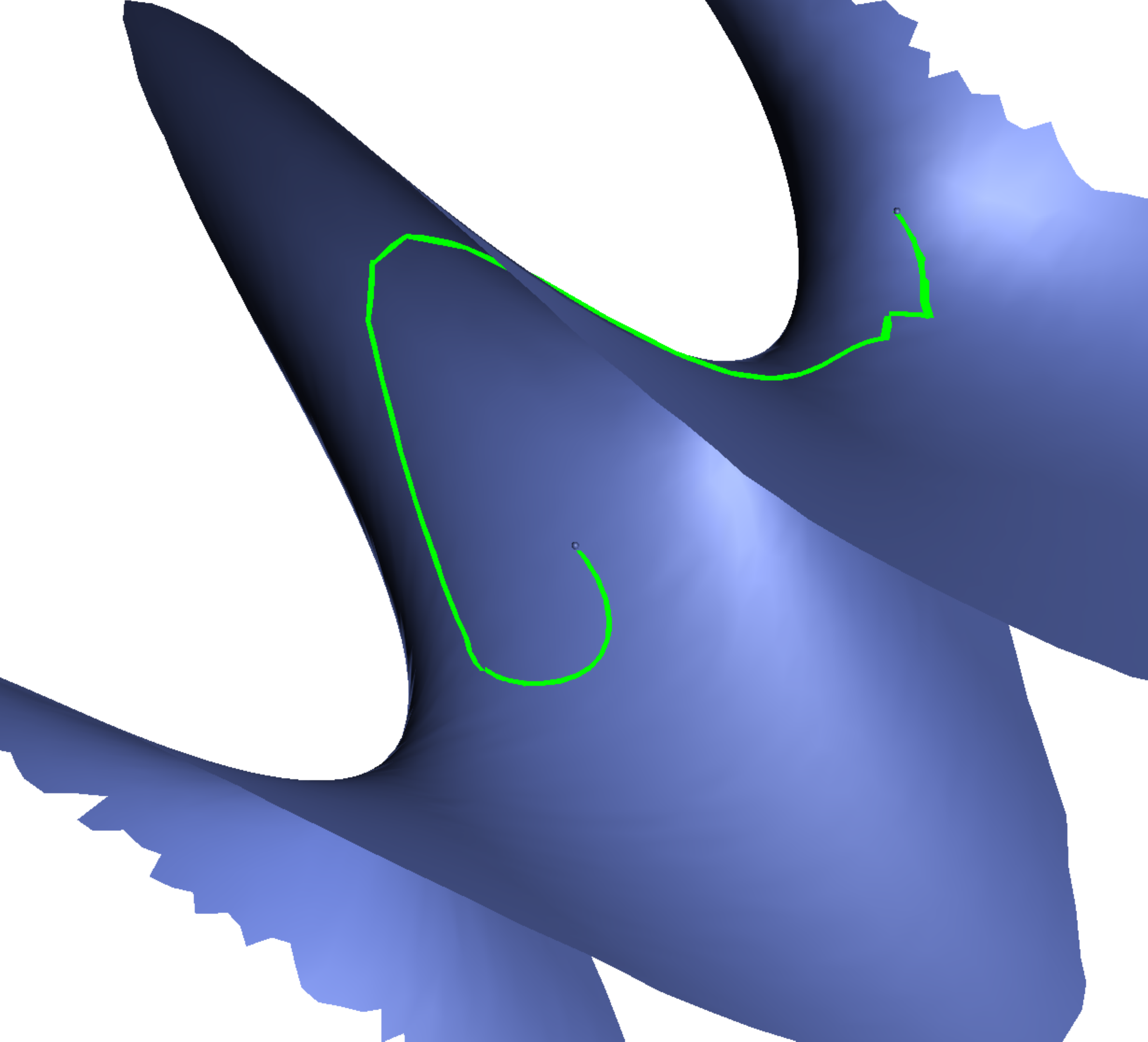} 
		\includegraphics[width=0.24\linewidth]{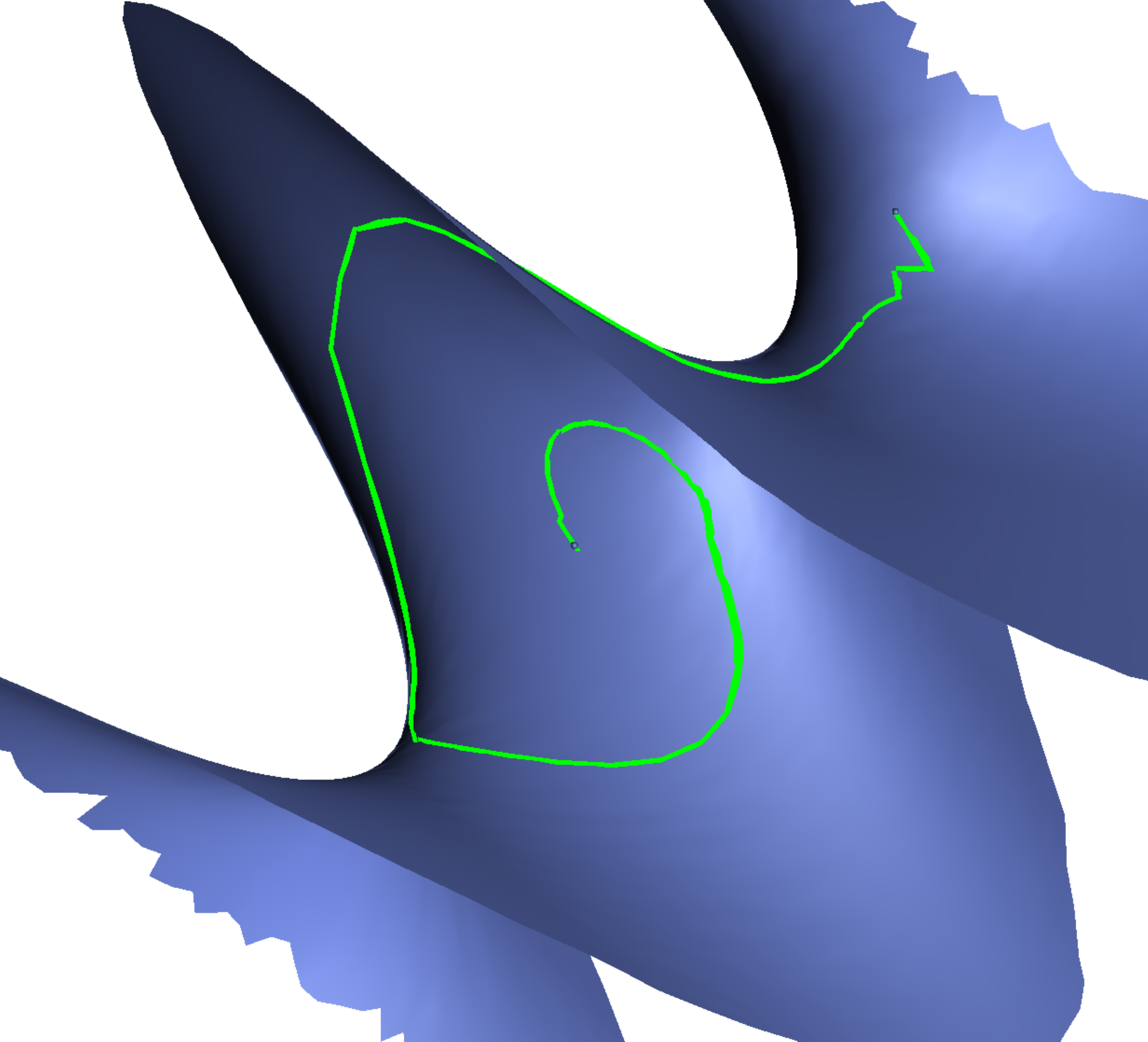}
		\includegraphics[width=0.24\linewidth]{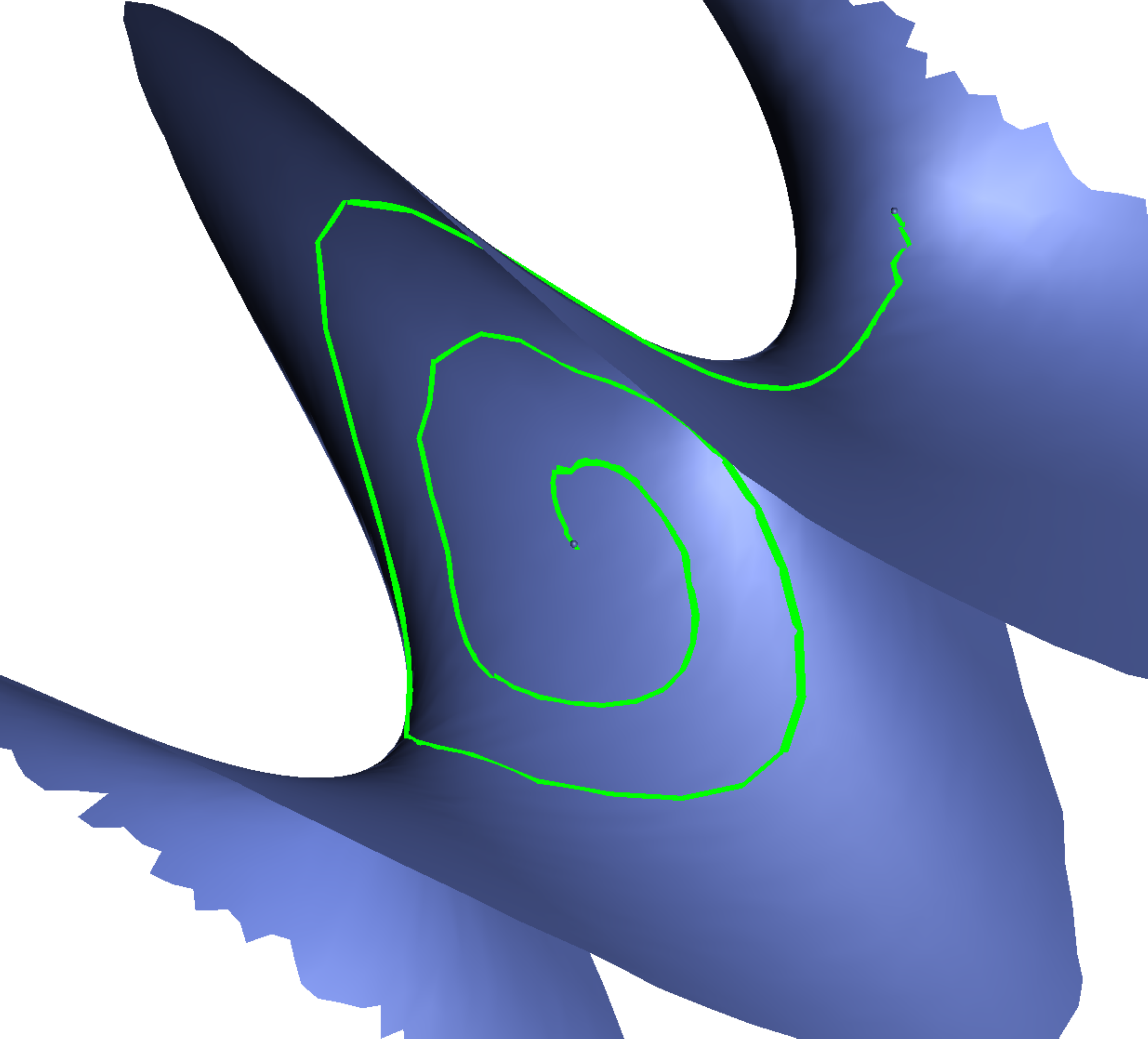} 
		\includegraphics[width=0.24\linewidth]{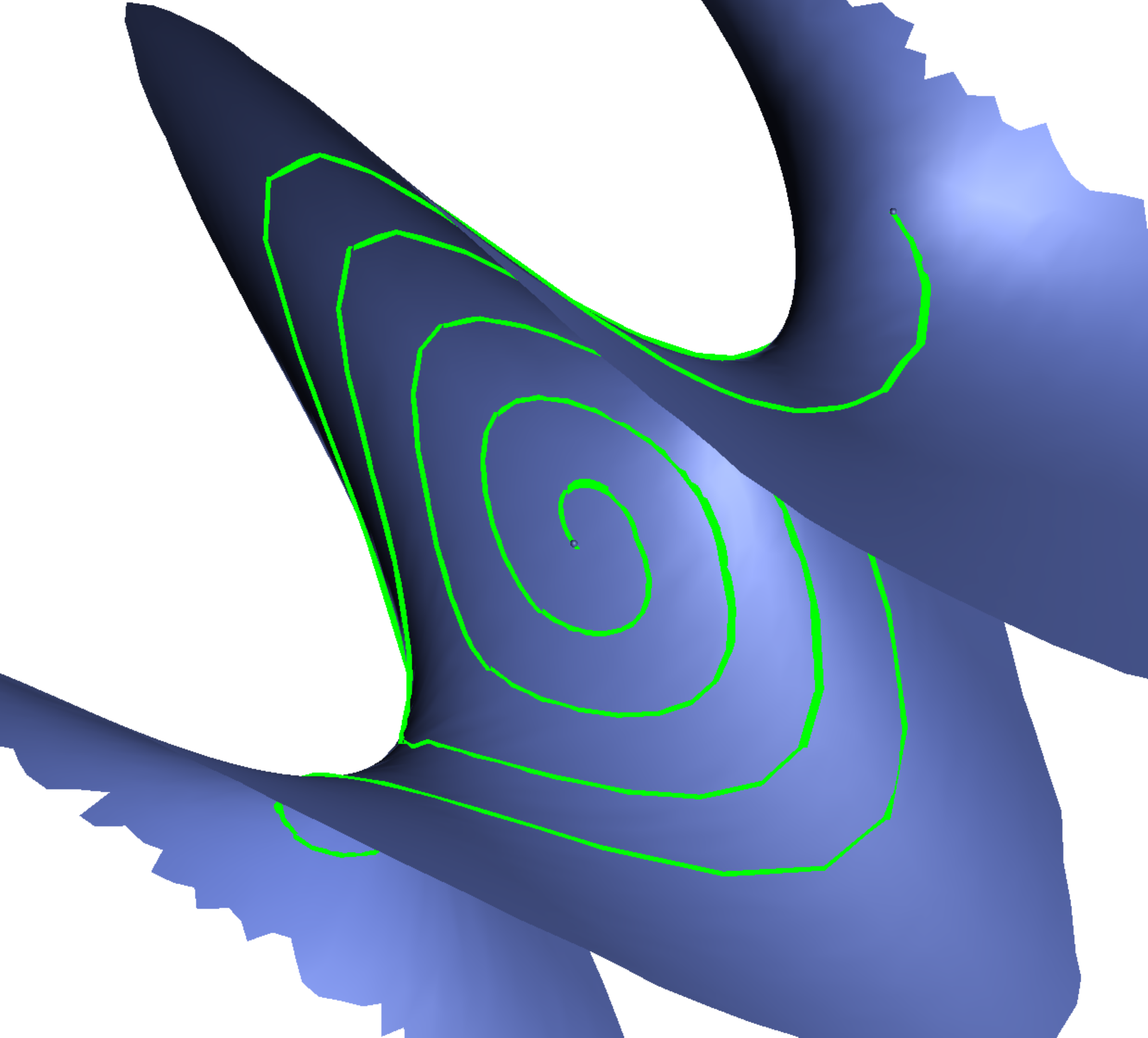}
		\caption{\label{fig:torque_max}From left to right, the state manifold (in blue), and the trajectory obtained (green) for a maximum torque $\tau_{max}$ of 16, 12, 8, and 4~[Nm]. }
	\end{center}
\end{figure*}

In this paper, relative joint angles are used to formulate
Eq.~\eqref{eq:phi}. As the three robots involve $n_q= 4$, $5$, and
$15$ joints, and each independent kinematic loop introduces $3$ or~$6$
constraints (depending on whether the robot is planar or spatial), the
dimensions of the C-space are $d_{\C}= 1$, $2$, and $3$, respectively.
For the formulation of the dynamic equations, the Euler-Lagrange
equations with multipliers are used. Friction forces can be easily
introduced in this formulation, but we have neglected them.

The robots respectively have $n_u=1$, $2$, and $3$ of their base
joints actuated, while the remaining joints are passive. Following
\citet{LaValle_IJRR01}, the set $\U$ is discretized in a way
reminiscent of bang-bang control schemes. Specifically, $\U$ will
contain all possible actions $\vr{u}$ in which only one actuator is
active at a time, with a torque that can be -$\tau_{max}$ or
$\tau_{max}$. Additionally, $\U$ also includes the action where no control is applied and the system simply drifts. As shown in Table~\ref{tab:param}, this results in sets
$\mathcal{U}$ of $3$,~$5$, and~$7$ actions, respectively. The table
also gives the $\tau_{max}$ values for each case.

In all test cases the parameters are set to
$t_m=0.1$, $\delta = 0.05$, $\rho_s=d_{\C}$, $\rho=\rho_s /2$,
$\cos(\alpha)=0.1$, and $\epsilon = 0.1$. The value $\beta$ is
problem-specific and given in Table~\ref{tab:param}. The table also
shows the performance statistics on an iMac with a Intel i7 processor
at 2.93 Ghz with 8 CPU cores, which are exploited to run
lines~\ref{alg:simulateAction} to~\ref{alg:simulateAction2} of
Algorithm~\ref{alg:extend} in parallel. The statistics include the
number of samples and charts, as well as the execution times in
seconds, all averaged over ten runs. The planner successfully
connected the start and goal states in all runs.

\begin{figure*}[t!]
	\begin{center}
		\includegraphics[width=0.15\linewidth]{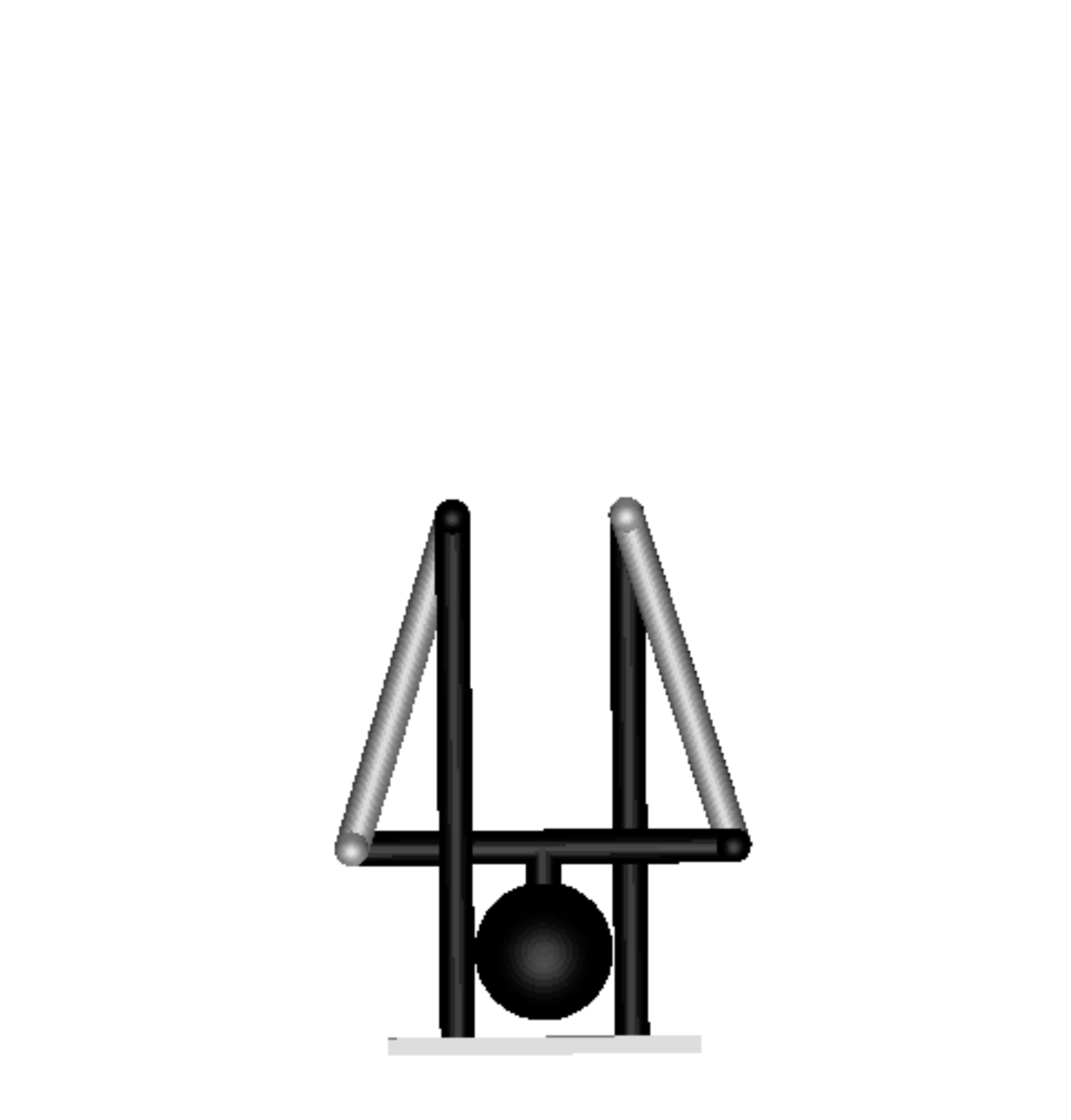} 
		\includegraphics[width=0.15\linewidth]{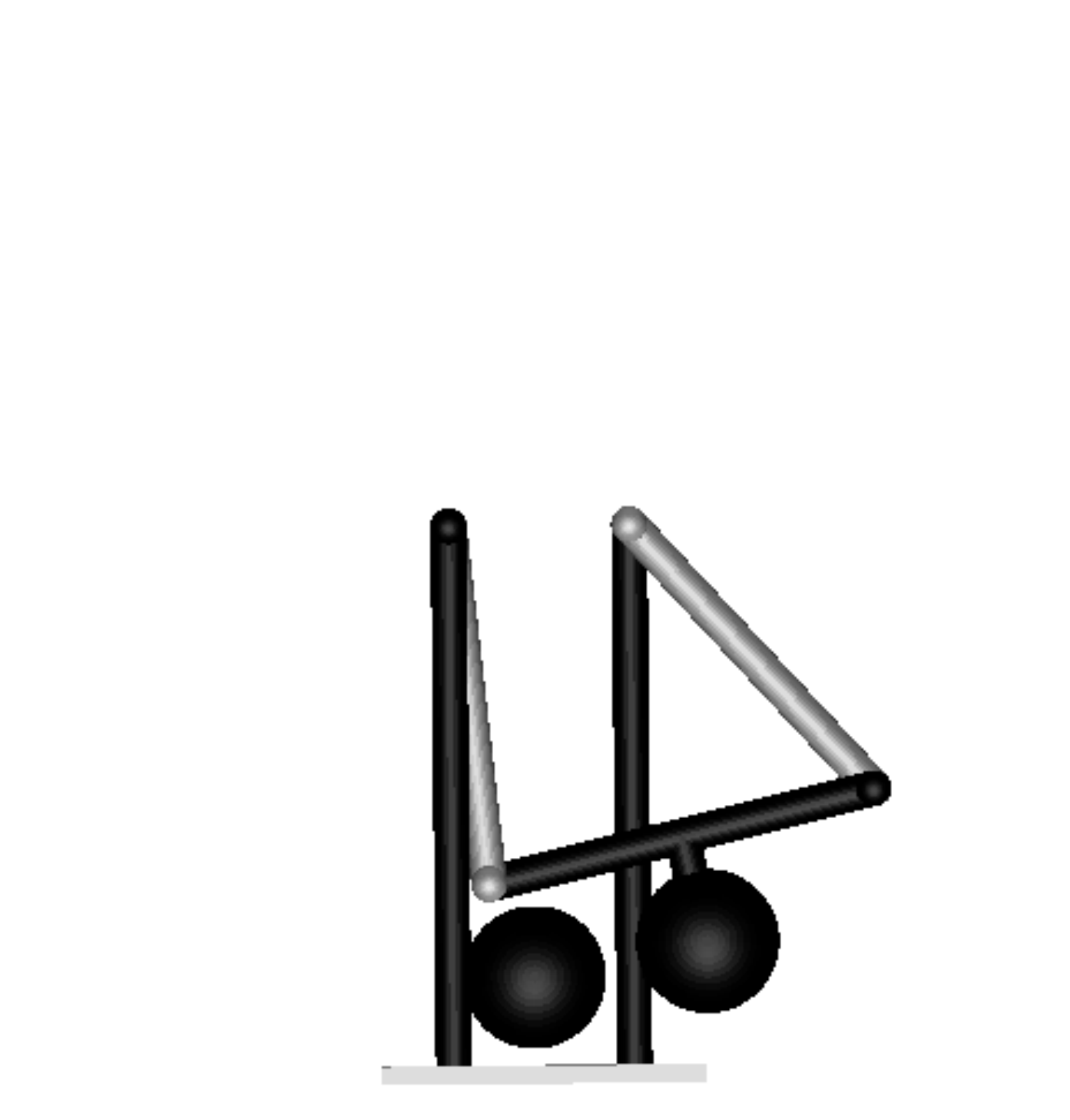} 
		\includegraphics[width=0.15\linewidth]{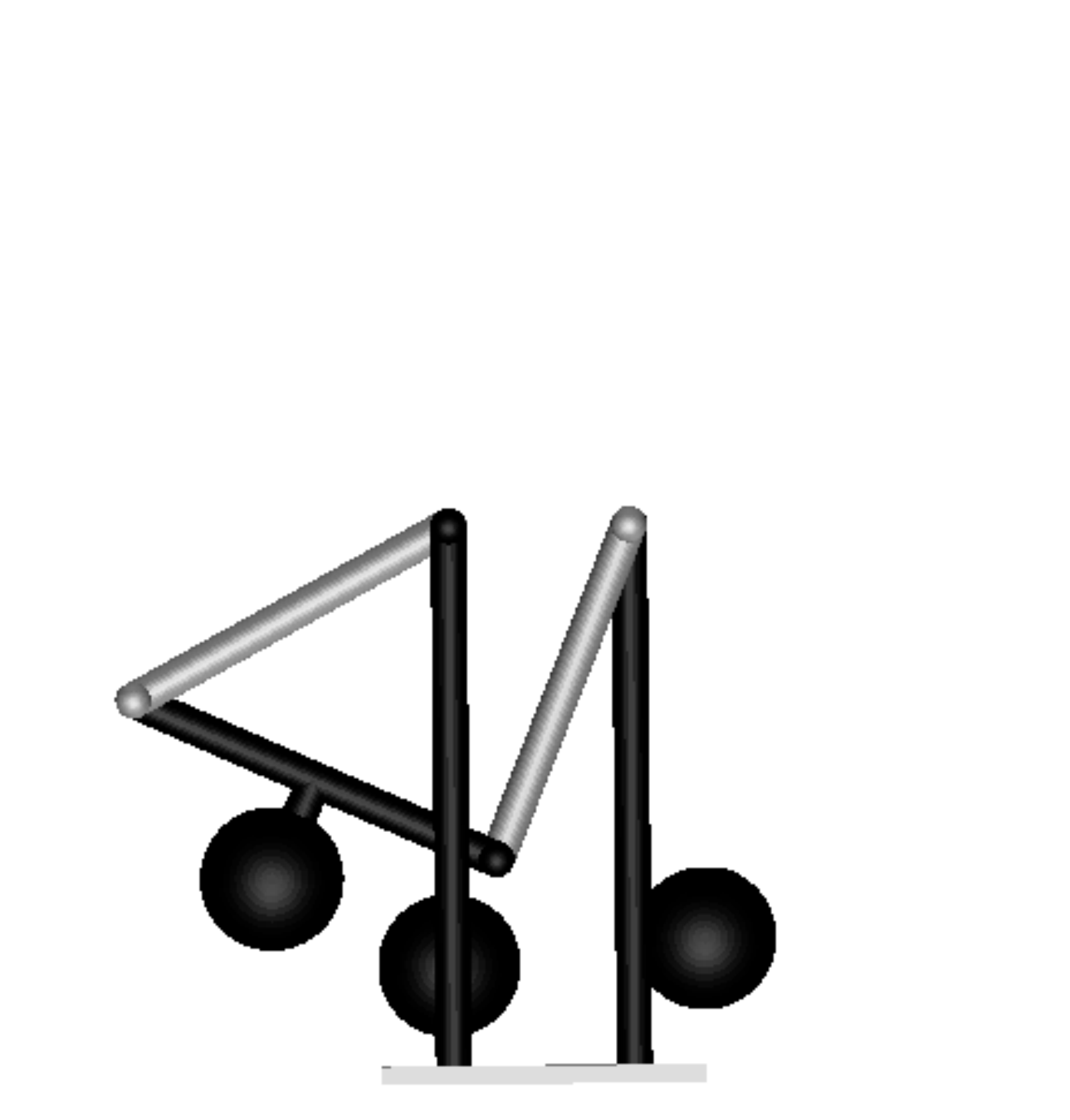} 
		\includegraphics[width=0.15\linewidth]{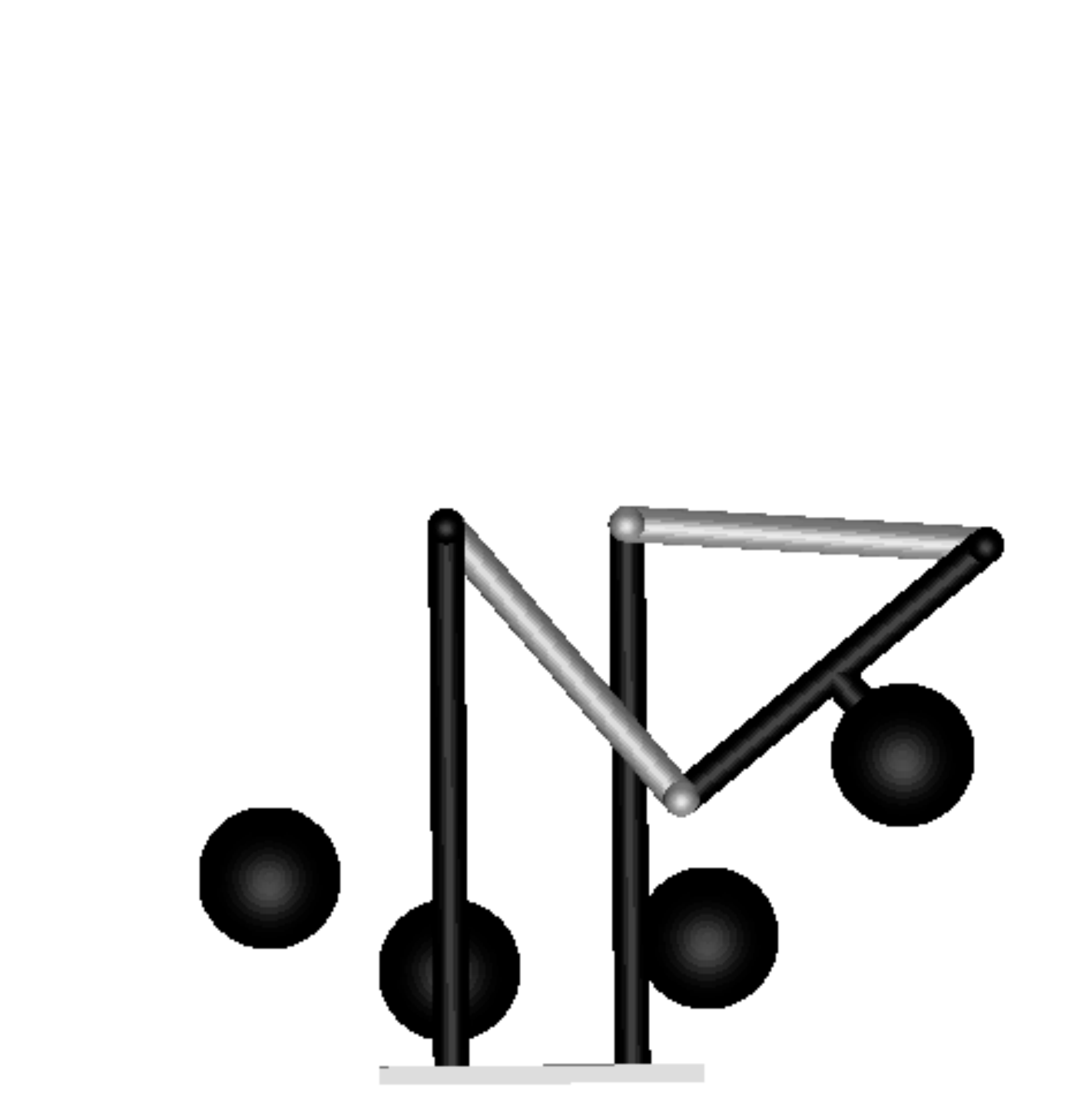}
		\includegraphics[width=0.15\linewidth]{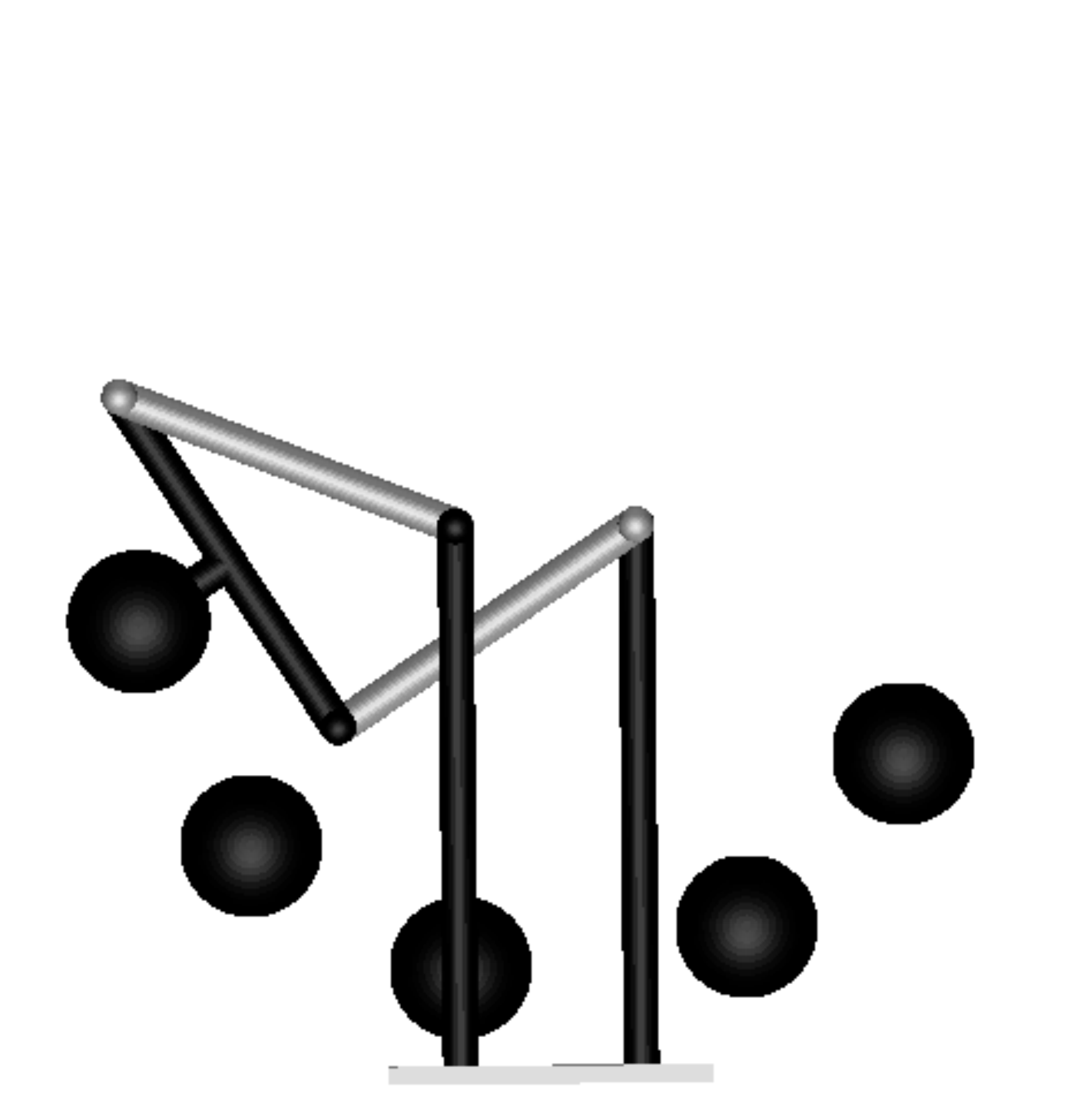}
		\includegraphics[width=0.15\linewidth]{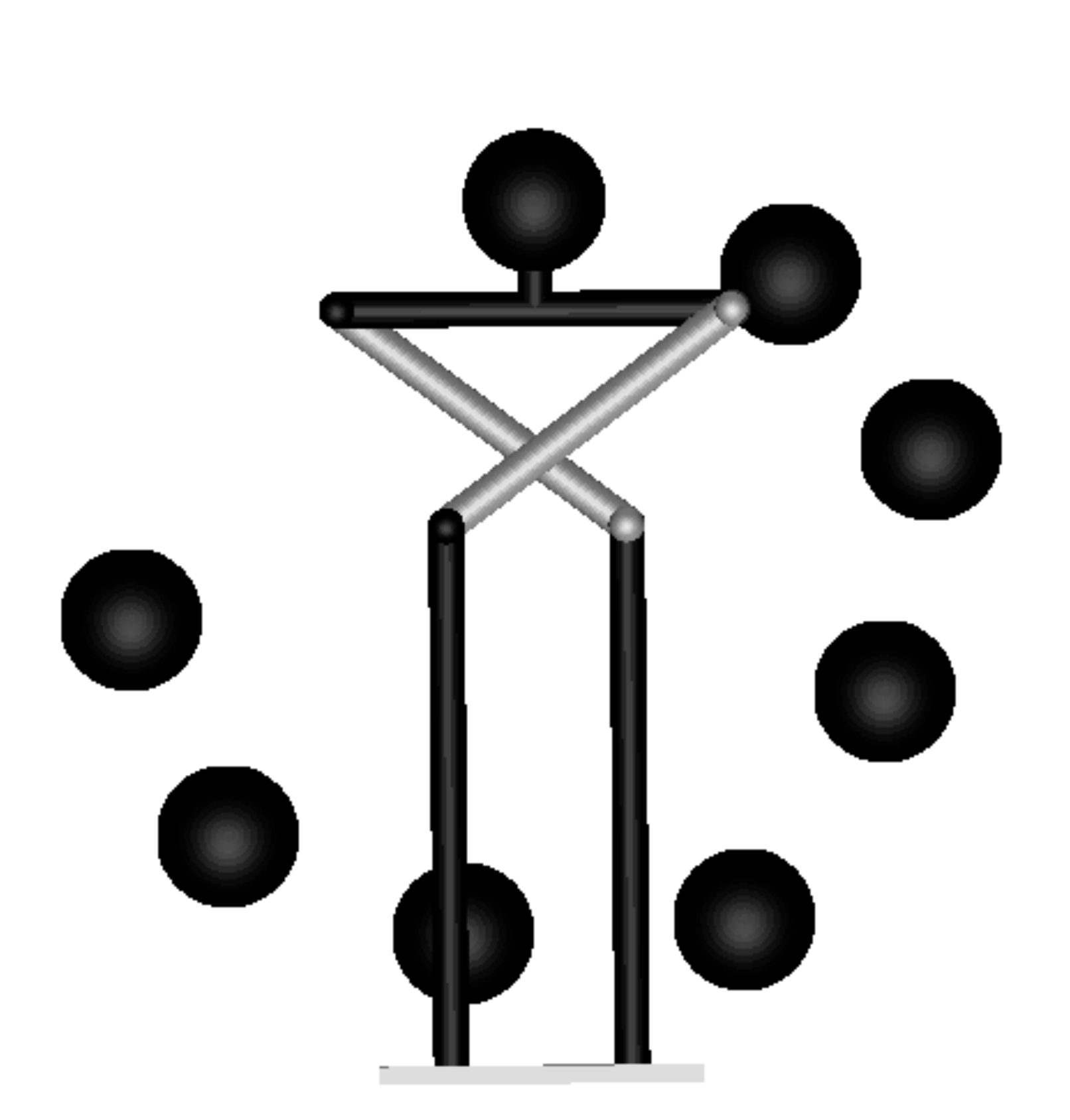}\\
		\includegraphics[width=0.19\linewidth]{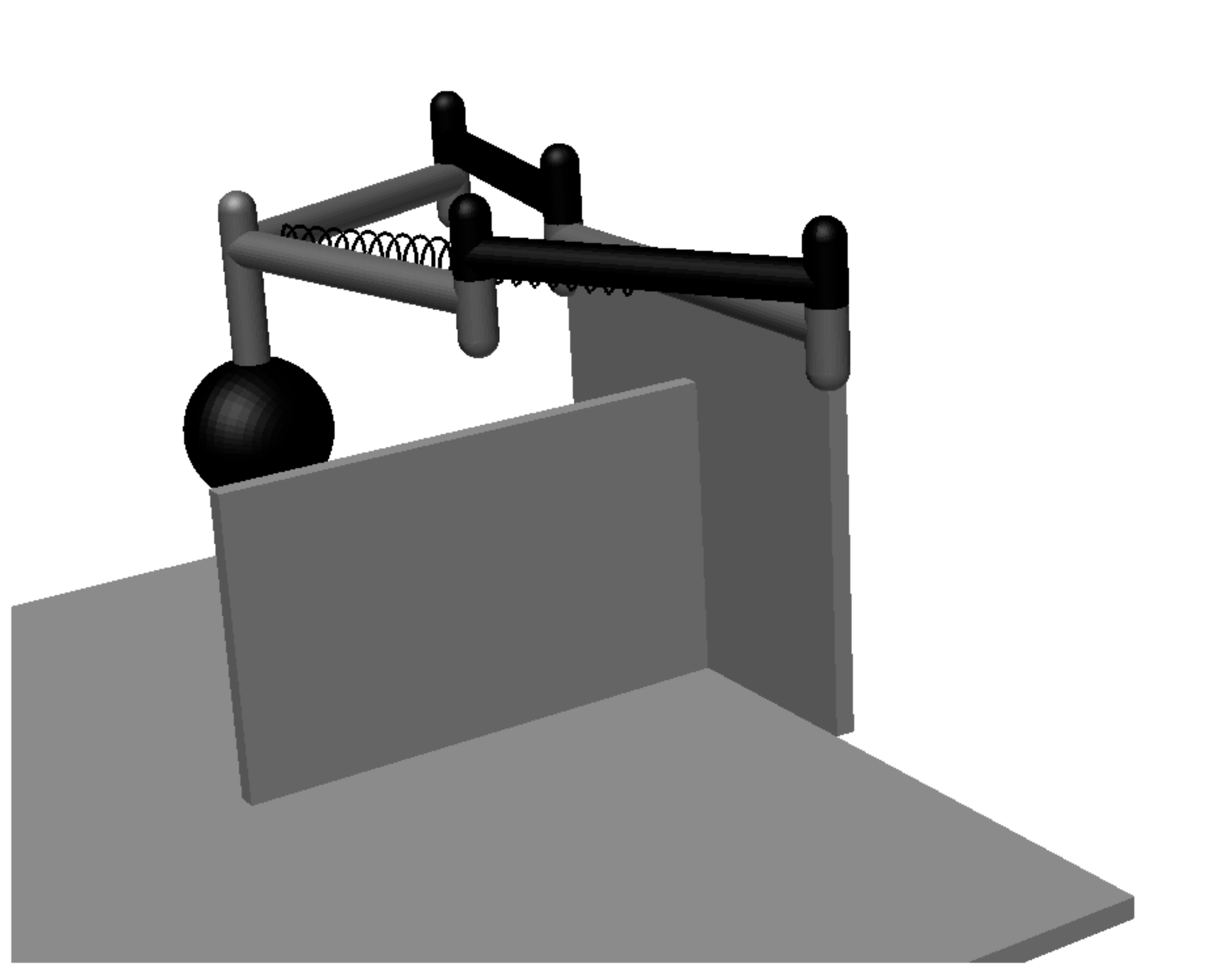} 
		\includegraphics[width=0.19\linewidth]{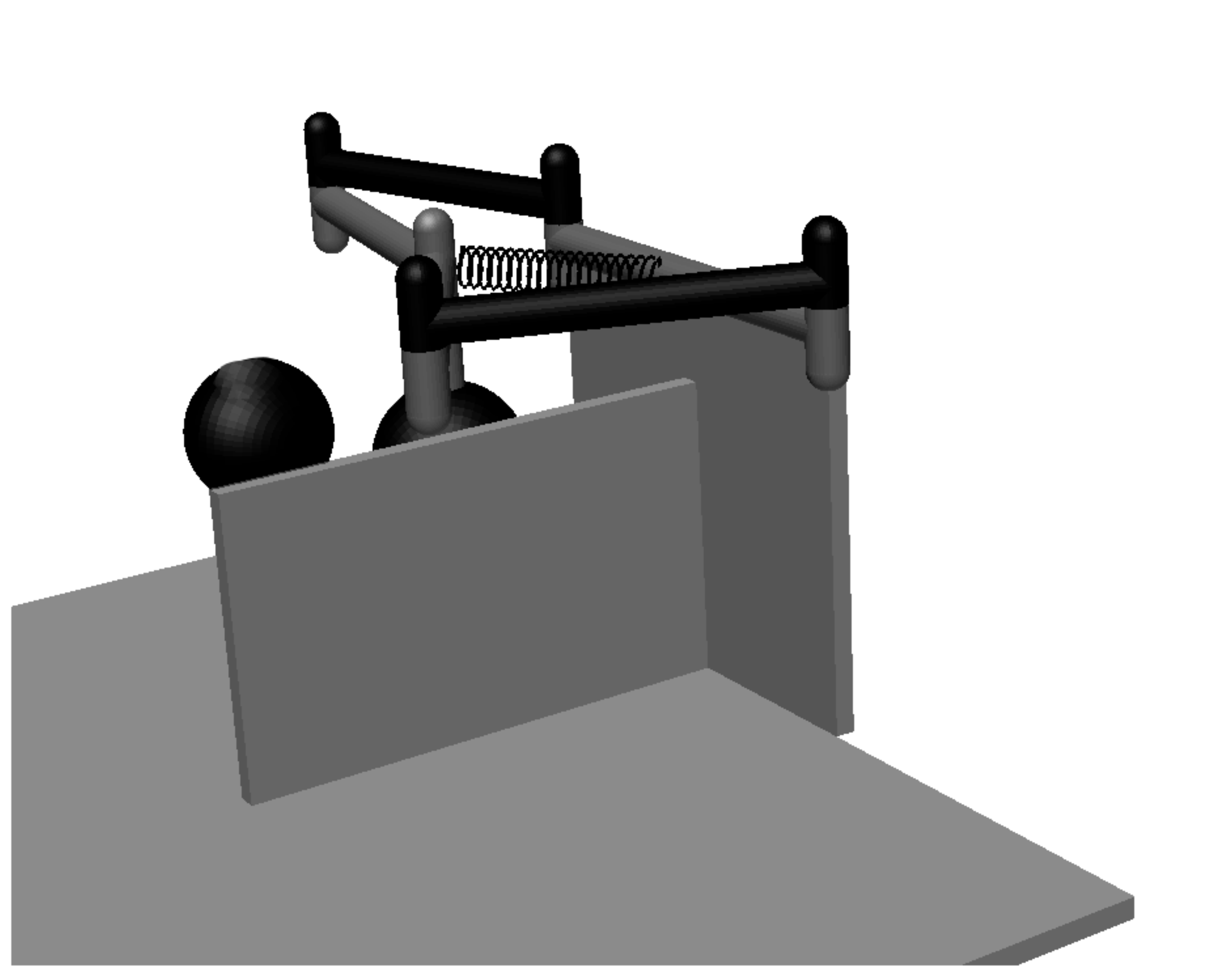} 
		\includegraphics[width=0.19\linewidth]{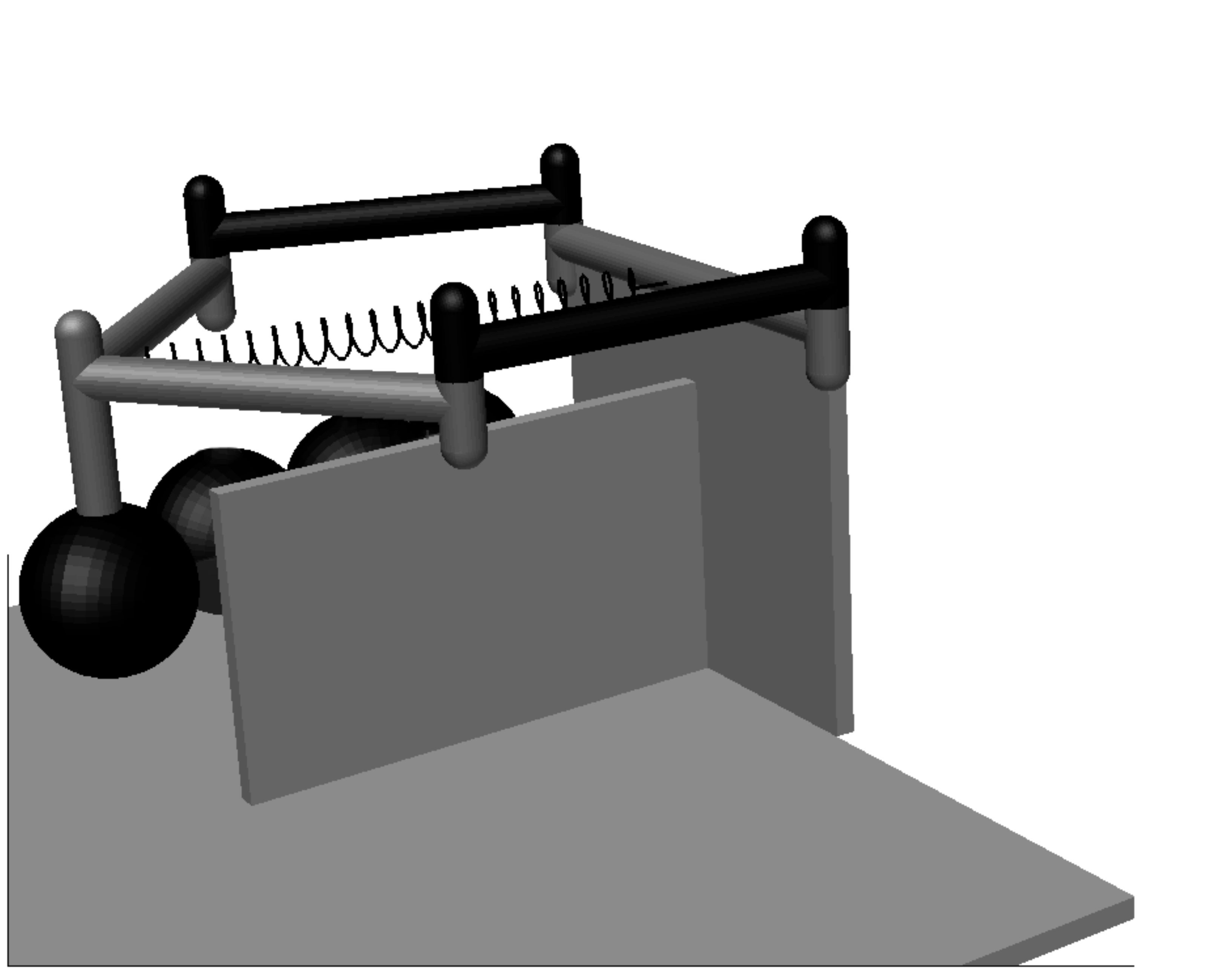} 
		\includegraphics[width=0.19\linewidth]{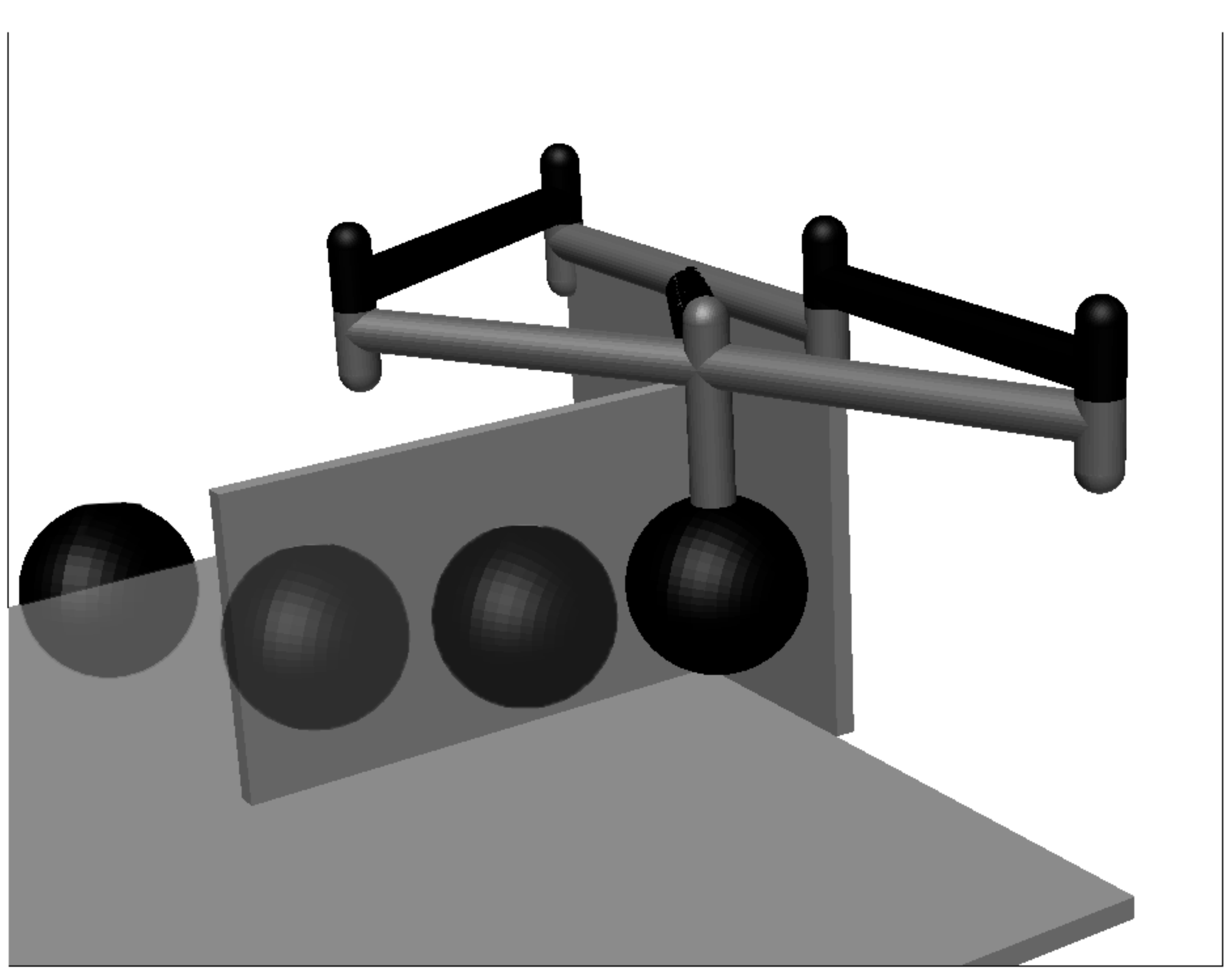}
		\includegraphics[width=0.19\linewidth]{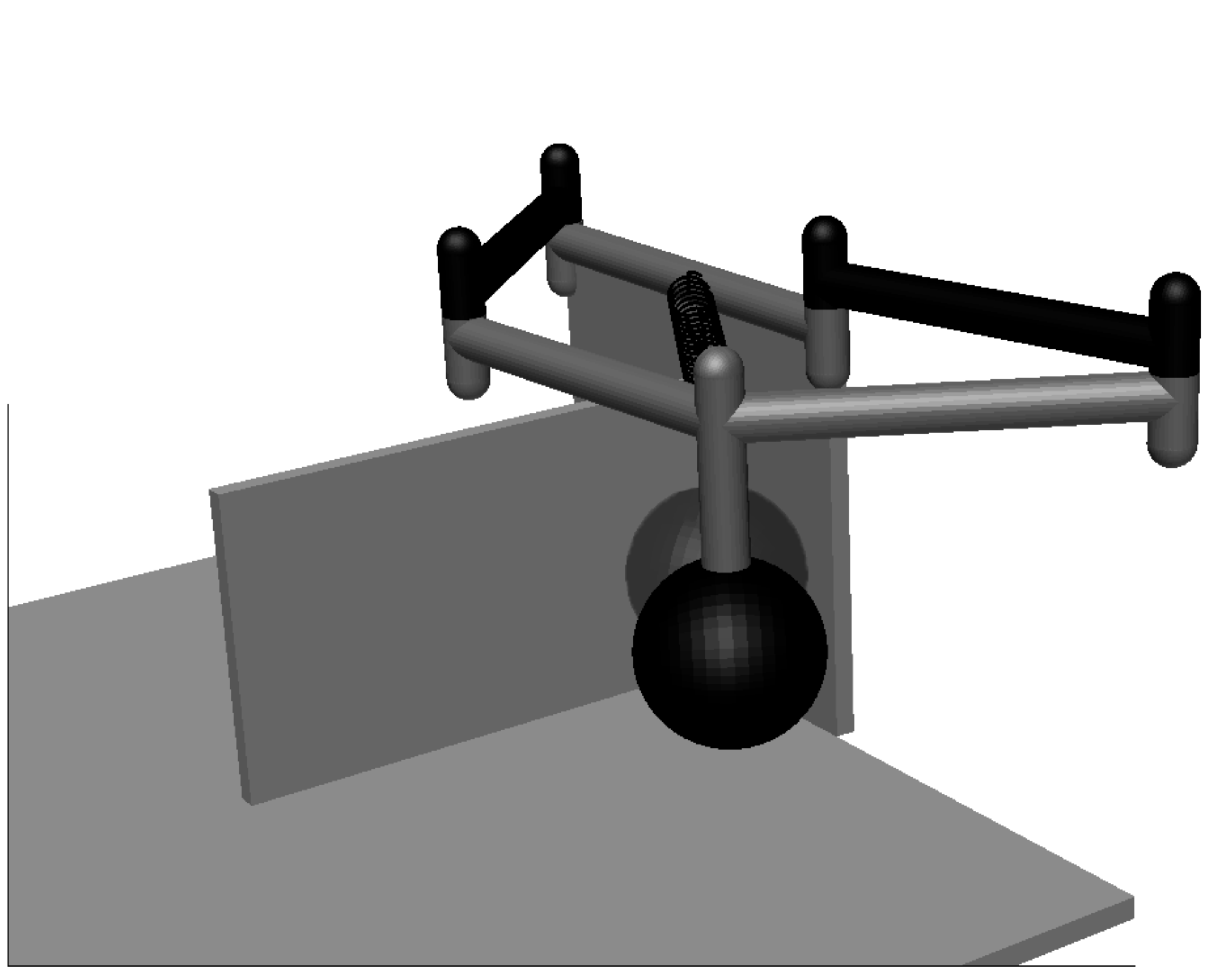}\\
		\includegraphics[width=0.32\linewidth]{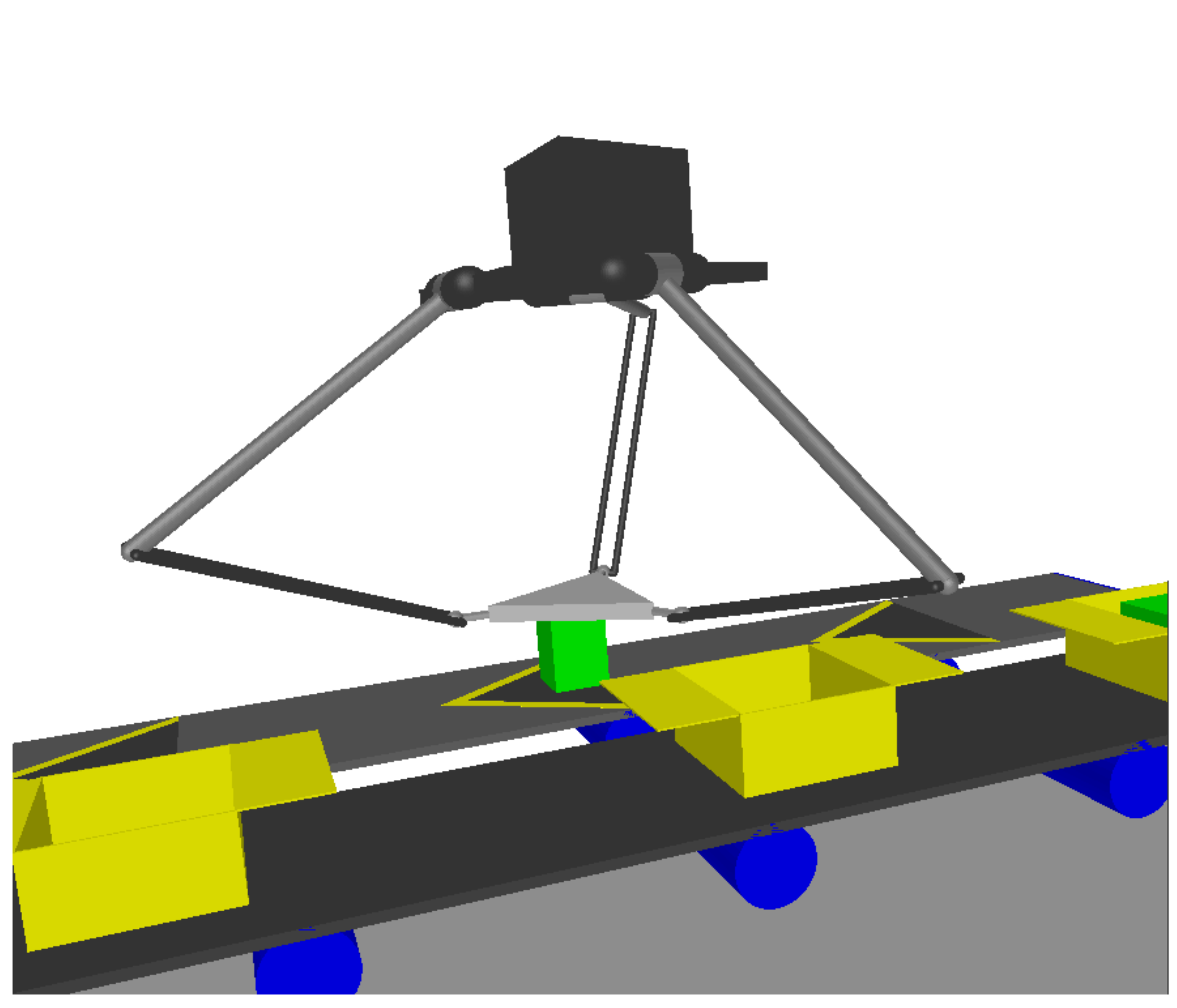} 
		\includegraphics[width=0.32\linewidth]{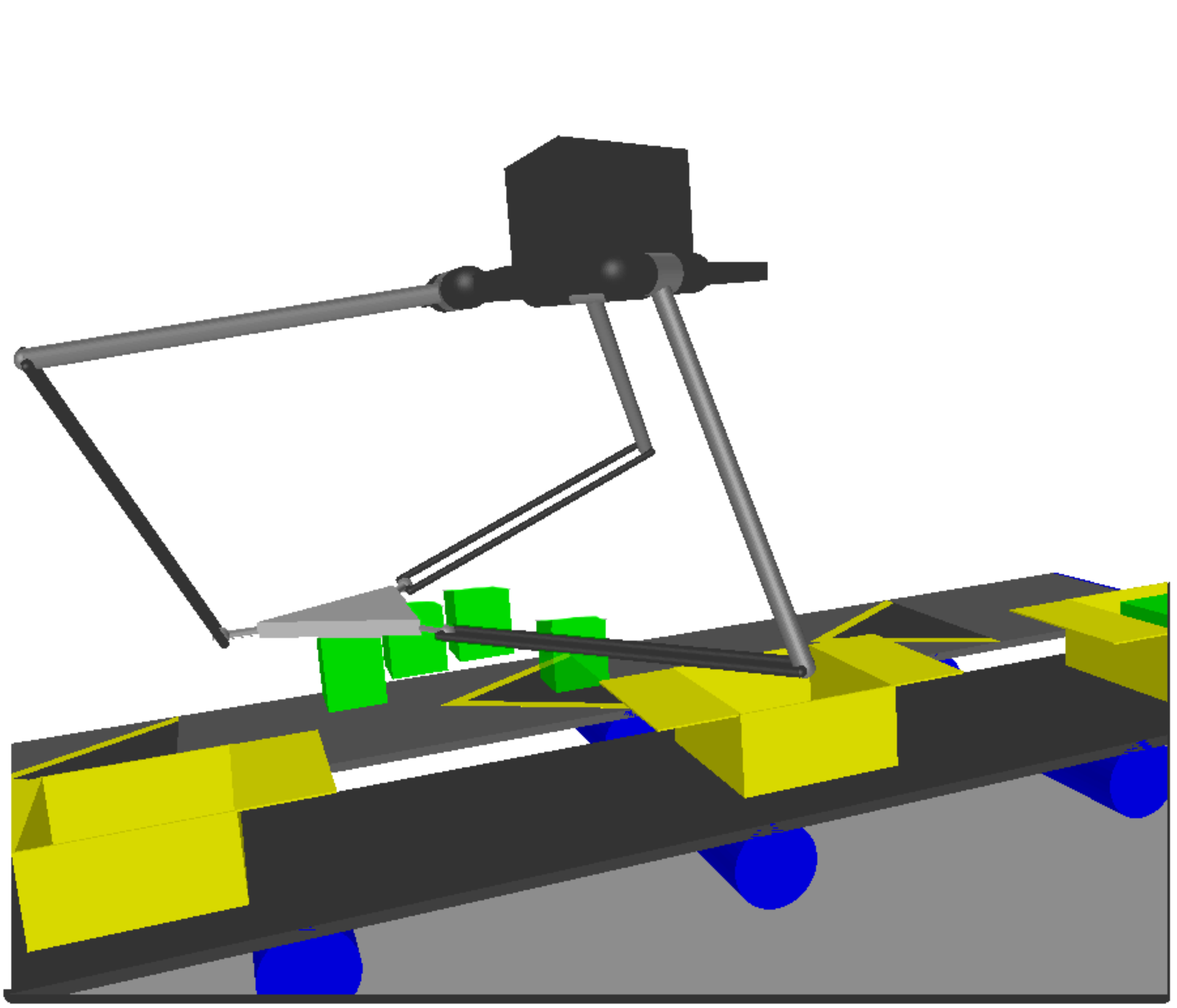} 
		\includegraphics[width=0.32\linewidth]{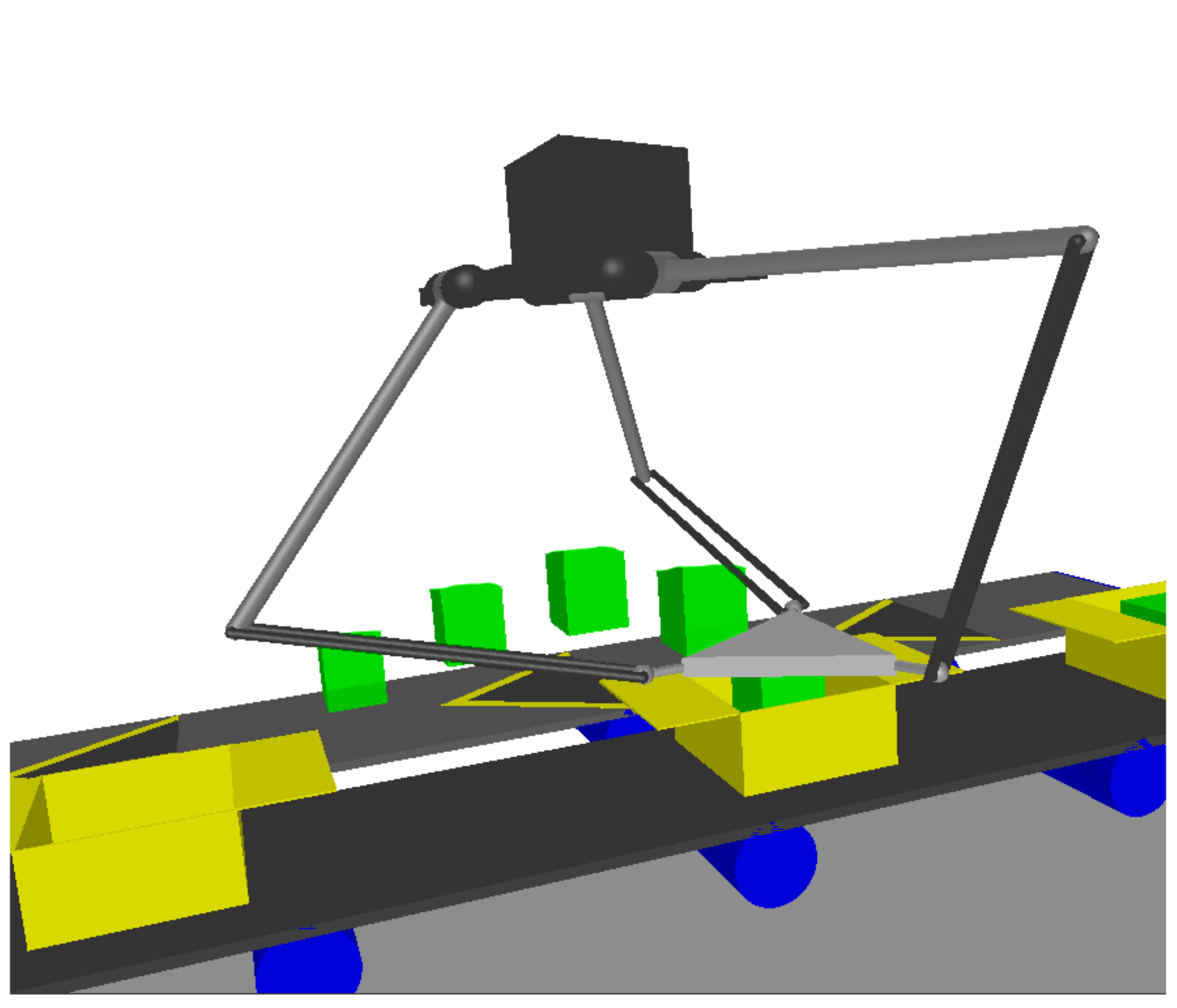} 
		\caption{\label{fig:trajectories} Snapshots of the trajectories obtained by the planner for the three test cases. Video animations can be seen in the multimedia files of this paper.}
	\end{center}
\end{figure*}

In the case of the four-bar mechanism, results are included for
decreasing values of $\tau_{max}$. As reflected in
Table~\ref{tab:param}, the lower the torque, the harder the planning
problem. The solution trajectory on the state-space manifold
(projected in one position and two velocity variables) can be seen in
Fig.~\ref{fig:torque_max} for the different values. Clearly, the number of
oscillations needed to reach the goal is successively higher. The
trajectory obtained for the most restricted case is shown in
Fig.~\ref{fig:trajectories}, top.

In the five-bars robot, although it only has one more link than the
previous robot, the planning problem is significantly more complex.
This is due to the narrow corridor created by the obstacle to
overcome. Moreover, the motors have a severely limited torque taking
into account the spring constant. In order to move the weight in such
conditions, the planner is forced to increase the momentum of the
payload before overpassing the obstacle, and to decrease it once it is
passed so as to reach the goal configuration with zero velocity
(Fig.~\ref{fig:trajectories},~middle). This increased complexity is
reflected in the number of samples and the execution time needed to
solve the problem. However, the number of charts is low, which shows
that the planner is not exploring new regions of the state space, but
trying to find a way through the narrow corridor.

Finally, the table gives the same statistics for the  problem on the
Delta robot. The execution time is higher due to the increased
complexity of the problem. The robot is spatial, it has a state space
of dimension $6$ in an ambient space of dimension $30$, and involves
more kinematic constraints than in the previous cases. Moreover, given
the velocity of the belt, the planner is forced to reduce the initial
momentum of the load before it can place it inside the box
(Fig.~\ref{fig:trajectories}, bottom).

\section{Conclusions}  \label{sec:conclusion}

This paper has proposed an RRT planner for dynamical systems with
implicitly defined state spaces. Dealing with such spaces presents two
major hurdles: the generation of random samples in the state space and
the driftless simulation over such space. We have seen that both
issues can be properly addressed by relying on local
parameterizations. The result is a planner that navigates the state
space manifold following the vector fields defined by the
dynamic constraints on such manifold. The presented experiments show
that the proposed method can successfully solve significantly complex
problems. In the current implementation, however, most of the time is
used in computing dynamics-related quantities. The use of specialized
dynamical simulation libraries may alleviate this
issue~\cite{TrepWeb,DartWeb}.

To scale to even more complex problems, several aspects of the
proposed RRT planner need to be improved. Probably the main issue is
the metric used to measure the distance between states. This is a
general issue of all sampling-based kinodynamic planners, but in our
context it is harder since the metric should not only consider the
vector fields defined by the dynamic constraints, but also the
curvature of the state space manifold defined by the kinematic
equations. Using a metric derived from geometric insights provided by
the kinodynamic constraints may result in significant performance
improvements. In a broader context, another aspect deserving attention
is the analysis of the proposed algorithm, in particular its
completeness and the optimality of the resulting trajectory. The
former should derive from the properties of the underlying planners.
With respect to the latter, locally optimal trajectories could be
obtained using the output of the planner to feed optimization
approaches~\cite{Posa_ICRA2016,Butler_WAFR2016}, or globally optimal
ones considering the trajectory cost into the planner
\cite{Hauser_TRO2016,Li_IJRR2016}. Finally, the relation of the
proposed approach with variational integration
methods~\cite{Marsden_ACTAMECH2001,Johnson_TRO2009} should also be
examined.

\section*{Acknowledgments}

This research has been partially funded by the Spanish Ministry of Economy under project DPI2014-57220-C2-2-P.



\bibliographystyle{plainnat}
\bibliography{references}

\begin{thebibliography}{54}
\providecommand{\natexlab}[1]{#1}
\providecommand{\url}[1]{\texttt{#1}}
\expandafter\ifx\csname urlstyle\endcsname\relax
  \providecommand{\doi}[1]{doi: #1}\else
  \providecommand{\doi}{doi: \begingroup \urlstyle{rm}\Url}\fi

\bibitem[Barraquand and Latombe(1991)]{Barraquand_ICRA1991}
J.~Barraquand and J.~C. Latombe.
\newblock \href{http://doi.org/10.1109/ROBOT.1991.131750}{Nonholonomic
  multibody mobile robots: controllability and motion planning in the presence
  of obstacles}.
\newblock In \emph{IEEE International Conference on Robotics and Automation},
  volume~3, pages 2328--2335, 1991.

\bibitem[Bauchau and Laulusa(2008)]{Bauchau_JCND2008}
O.~A. Bauchau and A.~Laulusa.
\newblock \href{http:/doi.org/10.1115/1.2803258}{Review of contemporary
  approaches for constraint enforcement in multibody systems}.
\newblock \emph{Journal of Computational and Nonlinear Dynamics}, 3\penalty0
  (1):\penalty0 011005, 2008.

\bibitem[Baumgarte(1972)]{Baumgarte_CMAME1972}
J.~Baumgarte.
\newblock \href{http://doi.org/10.1016/0045-7825(72)90018-7}{Stabilization of
  constraints and integrals of motion in dynamical systems}.
\newblock \emph{Computer Methods in Applied Mechanics and Engineering},
  1\penalty0 (1):\penalty0 1--16, 1972.

\bibitem[Berenson et~al.(2011)Berenson, Srinivasa, and
  Kuffner]{Berenson_IJRR11}
D.~Berenson, S.~Srinivasa, and J.~J. Kuffner.
\newblock \href{http://doi.org/10.1177/0278364910396389}{Task space regions: A
  framework for pose-constrained manipulation planning}.
\newblock \emph{International Journal of Robotics Research}, 30\penalty0
  (12):\penalty0 1435--1460, 2011.

\bibitem[Betts(2010)]{Betts_SIAM2010}
J.~T. Betts.
\newblock \emph{\href{http://doi.org/10.1137/1.9780898718577}{Practical Methods
  for Optimal Control and Estimation Using Nonlinear Programming}}.
\newblock SIAM Publications, 2010.

\bibitem[Blajer(2002)]{Blajer_MSD2002}
W.~Blajer.
\newblock \href{http://doi.org/10.1023/A:1015285428885}{Elimination of
  constraint violation and accuracy aspects in numerical simulation of
  multibody systems}.
\newblock \emph{Multibody System Dynamics}, 7\penalty0 (3):\penalty0 265--284,
  2002.

\bibitem[Bohigas et~al.(2013)Bohigas, Henderson, Ros, Manubens, and
  Porta]{Bohigas_TRO2013}
O.~Bohigas, M.~E. Henderson, L.~Ros, M.~Manubens, and J.~M. Porta.
\newblock \href{http://doi.org/10.1109/tro.2013.2260679}{Planning
  singularity-free paths on closed-chain manipulators}.
\newblock \emph{IEEE Transactions on Robotics}, 29\penalty0 (4):\penalty0
  888--898, 2013.

\bibitem[Bohigas et~al.(2016)Bohigas, Manubens, and Ros]{Bohigas_SPRINGER2016}
O.~Bohigas, M.~Manubens, and L.~Ros.
\newblock
  \emph{\href{http://www.springer.com/us/book/9783319329208}{Singularities of
  robot mechanisms: numerical computation and avoidance path planning}},
  volume~41 of \emph{Mechanisms and Machine Science}.
\newblock Springer, 2016.

\bibitem[Bohlin and Kavraki(2000)]{Bohlin_ICRA2000}
R.~Bohlin and L.~E. Kavraki.
\newblock \href{http://doi.org/10.1109/ROBOT.2000.844107}{Path planning using
  lazy PRM}.
\newblock In \emph{IEEE International Conference on Robotics and Automation},
  volume~1, pages 521--528, 2000.

\bibitem[Bourbonnais et~al.(2015)Bourbonnais, Bigras, and
  Bonev]{Bourbonnais_TMEC2015}
F.~Bourbonnais, P.~Bigras, and I.~A. Bonev.
\newblock \href{https://doi.org/10.1109/tmech.2014.2318999}{Minimum-time
  trajectory planning and control of a pick-and-place five-bar parallel robot}.
\newblock \emph{IEEE/ASME Transactions on Mechatronics}, 20\penalty0
  (2):\penalty0 740--749, 2015.

\bibitem[Braun and Goldfarb(2009)]{Braun_CMAME2009}
D.~J. Braun and M.~Goldfarb.
\newblock \href{http://doi.org/10.1016/j.cma.2009.05.013}{Eliminating
  constraint drift in the numerical simulation of constrained dynamical
  systems}.
\newblock \emph{Computer Methods in Applied Mechanics and Engineering},
  198\penalty0 (37–40):\penalty0 3151--3160, 2009.

\bibitem[Butler et~al.(2016)Butler, Moll, and Kavraki]{Butler_WAFR2016}
S.~D. Butler, M.~Moll, and L.~E. Kavraki.
\newblock \href{http://www.wafr.org/papers/WAFR_2016_paper_81.pdf}{A general
  algorithm for time-optimal trajectory generation subject to minimum and
  maximum constraints}.
\newblock In \emph{Workshop on the Algorithmic Foundations of Robotics}, 2016.

\bibitem[Canny(1988)]{Canny_STC1988}
J.~Canny.
\newblock \href{http://doi.acm.org/10.1145/62212.62257}{Some algebraic and
  geometric computations in {PSPACE}}.
\newblock In \emph{ACM Symposium on Theory of Computing}, pages 460--467, 1988.

\bibitem[Cheng(2005)]{Cheng_PhD2005}
P.~Cheng.
\newblock \emph{\href{http://hdl.handle.net/2142/11080}{Sampling-based motion
  planning with differential constraints}}.
\newblock Phd dissertation, Department of Computer Science, University of
  Illinois, 2005.

\bibitem[Cheng and LaValle(2001)]{Cheng_IROS2001}
P.~Cheng and S.~M. LaValle.
\newblock \href{http://doi.org/10.1109/IROS.2001.973334}{Reducing metric
  sensitivity in randomized trajectory design}.
\newblock In \emph{IEEE/RSJ International Conference on Intelligent Robots and
  Systems}, volume~1, pages 43--48, 2001.

\bibitem[Cheng and LaValle(2002)]{Cheng_ICRA2002}
P.~Cheng and S.~M. LaValle.
\newblock \href{http://doi.org/10.1109/ROBOT.2002.1013372}{Resolution complete
  rapidly-exploring random trees}.
\newblock In \emph{IEEE International Conference on Robotics and Automation},
  volume~1, pages 267--272, 2002.

\bibitem[Cherif(1999)]{Cherif_ICRA1999}
M.~Cherif.
\newblock \href{http://doi.org/10.1109/ROBOT.1999.769998}{Kinodynamic motion
  planning for all-terrain wheeled vehicles}.
\newblock In \emph{IEEE International Conference on Robotics and Automation},
  volume~1, pages 317--322, 1999.

\bibitem[Choset et~al.(2005)Choset, Lynch, Hutchinson, Kantor, Burgard,
  Kavraki, and Thrun]{Choset_05}
H.~Choset, K.~M. Lynch, S.~Hutchinson, G.~A. Kantor, W.~Burgard, L.~E. Kavraki,
  and S.~Thrun.
\newblock
  \emph{\href{http://mitpress.mit.edu/books/principles-robot-motion}{Principles
  of Robot Motion: Theory, Algorithms, and Implementations}}.
\newblock Intelligent Robotics and Autonomous Agents. MIT Press, 2005.

\bibitem[Donald et~al.(1993)Donald, Xavier, Canny, and Reif]{Donald_JACM1993}
B.~Donald, P.~Xavier, J.~Canny, and J.~Reif.
\newblock \href{http://doi.org/10.1145/174147.174150}{Kinodynamic motion
  planning}.
\newblock \emph{Journal of the ACM}, 40\penalty0 (5):\penalty0 1048–--1066,
  1993.

\bibitem[Dubowsky and Shiller(1985)]{Dubowsky_ROMANSY1985}
S.~Dubowsky and Z.~Shiller.
\newblock \href{http://doi.org/10.1007/978-1-4615-9882-4_15}{Optimal dynamic
  trajectories for robotic manipulators}.
\newblock In \emph{IFToMM Symposium on Robot Design, Dynamics and Control},
  pages 133--143, 1985.

\bibitem[{Georgia Tech and Carnegie Mellon University}(2017)]{DartWeb}
{Georgia Tech and Carnegie Mellon University}.
\newblock {DART}: Dynamic animation and robotics toolkit.
\newblock \href{https://dartsim.github.io}{https://dartsim.github.io}, 2017.

\bibitem[Hairer(2001)]{Hairer_NM2001}
E.~Hairer.
\newblock \href{http://doi.org/10.1023/A:1021989212020}{Geometric integration
  of ordinary differential equations on manifolds}.
\newblock \emph{BIT Numerical Mathematics}, 41\penalty0 (5):\penalty0
  996--1007, 2001.

\bibitem[Hairer et~al.(2006)Hairer, Lubich, and Wanner]{Hairer_SPRINGER2006}
E.~Hairer, C.~Lubich, and G.~Wanner.
\newblock \href{http://doi.org/10.1007/978-3-662-05018-7}{Geometric numerical
  integration: structure-preserving algorithms for ordinary differential
  equations}, 2006.

\bibitem[Hauser and Zhou(2016)]{Hauser_TRO2016}
K.~Hauser and Y.~Zhou.
\newblock \href{http://doi.org/10.1109/TRO.2016.2602363}{Asymptotically optimal
  planning by feasible kinodynamic planning in a state-cost space}.
\newblock \emph{IEEE Transactions on Robotics}, 32\penalty0 (6):\penalty0
  1431--1443, 2016.

\bibitem[Henderson(2002)]{Henderson_BC02}
M.~E. Henderson.
\newblock \href{http://doi.org/10.1142/S0218127402004498}{Multiple parameter
  continuation: computing implicitly defined k-manifolds}.
\newblock \emph{International Journal of Bifurcation and Chaos}, 12\penalty0
  (3):\penalty0 451--476, 2002.

\bibitem[Jaillet and Porta(2013)]{Jaillet_TRO2013}
L.~Jaillet and J.~M. Porta.
\newblock \href{http://doi.org/10.1109/TRO.2012.2222272}{Path planning under
  kinematic constraints by rapidly exploring manifolds}.
\newblock \emph{IEEE Transactions on Robotics}, 29\penalty0 (1):\penalty0
  105--117, 2013.

\bibitem[Jaillet et~al.(2011)Jaillet, Hoffman, van~den Berg, Abbeel, Porta, and
  Goldberg]{Jaillet_IROS2011}
L.~Jaillet, J.~Hoffman, J.~van~den Berg, P.~Abbeel, J.~M. Porta, and
  K.~Goldberg.
\newblock \href{http://doi.org/10.1109/IROS.2011.6094802}{EG-RRT:
  Environment-guided random trees for kinodynamic motion planning with
  uncertainty and obstacles}.
\newblock In \emph{IEEE/RSJ International Conference on Intelligent Robots and
  Systems}, pages 2646--2652, 2011.

\bibitem[Johnson and Murphey(2009)]{Johnson_TRO2009}
E.~R. Johnson and T.~D. Murphey.
\newblock \href{http://doi.org/10.1109/tro.2009.2032955}{Scalable variational
  integrators for constrained mechanical systems in generalized coordinates}.
\newblock \emph{{IEEE} Transactions on Robotics}, 25\penalty0 (6):\penalty0
  1249--1261, 2009.

\bibitem[Kalakrishnan et~al.(2011)Kalakrishnan, Chitta, Theodorou, Pastor, and
  Schaal]{Kalakrishnan_2011}
M.~Kalakrishnan, S.~Chitta, E.~Theodorou, P.~Pastor, and S.~Schaal.
\newblock \href{http://doi.org/10.1109/ICRA.2011.5980280}{STOMP: Stochastic
  trajectory optimization for motion planning}.
\newblock In \emph{2011 IEEE International Conference on Robotics and
  Automation}, pages 4569--4574, 2011.

\bibitem[Kalisiak(2008)]{Kalisiak_PhD2008}
M.~Kalisiak.
\newblock \emph{\href{http://hdl.handle.net/1807/11215}{Toward more efficient
  motion planning with differential constraints}}.
\newblock PhD thesis, Computer Science,University of Toronto, 2008.

\bibitem[Kavraki et~al.(1996)Kavraki, Svestka, Latombe, and
  Overmars]{Kavraki_TRA1996}
L.~E. Kavraki, P.~Svestka, J.-C. Latombe, and M.~H. Overmars.
\newblock \href{http://doi.org/10.1109/70.508439}{Probabilistic roadmaps for
  path planning in high-dimensional configuration spaces}.
\newblock \emph{IEEE Transactions on Robotics and Automation}, 12\penalty0
  (4):\penalty0 566--580, 1996.

\bibitem[Ladd and Kavraki(2005)]{Ladd_WAFR2005}
A.~M. Ladd and L.~E. Kavraki.
\newblock \href{http://doi.org/10.1007/10991541_21}{Fast tree-based exploration
  of state space for robots with dynamics}.
\newblock In \emph{Algorithmic Foundations of Robotics VI}, pages 297--312,
  2005.

\bibitem[Latombe(1991)]{Labombe_91}
J.-C. Latombe.
\newblock \emph{\href{http://doi.org/10.1007\%2F978-1-4615-4022-9}{Robot Motion
  Planning}}.
\newblock Engineering and Computer Science. Springer, 1991.

\bibitem[Laulusa and Bauchau(2008)]{Laulusa_JCND2008}
A.~Laulusa and O.~A. Bauchau.
\newblock \href{http:/doi.org/10.1115/1.2803257}{Review of classical approaches
  for constraint enforcement in multibody systems}.
\newblock \emph{Journal of Computational and Nonlinear Dynamics}, 3\penalty0
  (1):\penalty0 011004, 2008.

\bibitem[LaValle(2006)]{Lavalle_06}
S.~M. LaValle.
\newblock \emph{\href{http://planning.cs.uiuc.edu/}{Planning Algorithms}}.
\newblock Cambridge University Press, New York, 2006.

\bibitem[LaValle and Kuffner(2001)]{LaValle_IJRR01}
S.~M. LaValle and J.~J. Kuffner.
\newblock \href{http://doi.org/10.1177/02783640122067453}{Randomized
  kinodynamic planning}.
\newblock \emph{International Journal of Robotics Research}, 20\penalty0
  (5):\penalty0 378--400, 2001.

\bibitem[Lee(2001)]{Lee_2001}
J.~M. Lee.
\newblock \emph{\href{http://doi.org/10.1007/978-0-387-21752-9}{Introduction to
  Smooth Manifolds}}.
\newblock Springer, 2001.

\bibitem[Li et~al.(2016)Li, Littlefield, and Bekris]{Li_IJRR2016}
Y.~Li, Z.~Littlefield, and K.~E. Bekris.
\newblock \href{http://doi.org/10.1177/0278364915614386}{Asymptotically optimal
  sampling-based kinodynamic planning}.
\newblock \emph{The International Journal of Robotics Research}, 35\penalty0
  (5):\penalty0 528--564, 2016.

\bibitem[Lynch and Mason(1996)]{Lynch_IJRR1996}
K.~M. Lynch and M.~T. Mason.
\newblock \href{http://doi.org/10.1177/027836499601500602}{Stable pushing:
  mechanics, controllability, and planning}.
\newblock \emph{The International Journal of Robotics Research}, 15\penalty0
  (6):\penalty0 533--556, 1996.

\bibitem[Marsden and West(2001)]{Marsden_ACTAMECH2001}
J.~E. Marsden and M.~West.
\newblock \href{http://doi.org/10.1017/s096249290100006x}{Discrete mechanics
  and variational integrators}.
\newblock \emph{Acta Numerica 2001}, 10:\penalty0 357--514, 2001.

\bibitem[Petzold(1992)]{Petzold_PDNP1992}
L.~R. Petzold.
\newblock \href{http://doi.org/10.1016/0167-2789(92)90243-G}{Numerical solution
  of differential-algebraic equations in mechanical systems simulation}.
\newblock \emph{Physica D: Nonlinear Phenomena}, 60\penalty0 (1--4):\penalty0
  269--279, 1992.

\bibitem[Pham et~al.(2017)Pham, Caron, Lertkultanon, and
  Nakamura]{Pham_IJRR2017}
Q.-C. Pham, S.~Caron, P.~Lertkultanon, and Y.~Nakamura.
\newblock \href{http://doi.org/10.1177/0278364916675419}{Admissible velocity
  propagation: Beyond quasi-static path planning for high-dimensional robots}.
\newblock \emph{The International Journal of Robotics Research}, 36\penalty0
  (1):\penalty0 44--67, 2017.

\bibitem[Plaku et~al.(2007)Plaku, Kavraki, and Vardi]{Plaku_RSS07}
E.~Plaku, L.~Kavraki, and M.~Vardi.
\newblock \href{http://doi.org/10.15607/RSS.2007.III.040}{Discrete search
  leading continuous exploration for kinodynamic motion planning}.
\newblock In \emph{Proceedings of Robotics: Science and Systems}, 2007.

\bibitem[Porta et~al.(2012)Porta, Jaillet, and Bohigas]{Porta_IJRR2012}
J.~M. Porta, L.~Jaillet, and O.~Bohigas.
\newblock \href{http://doi.org/10.1177/0278364911432324}{Randomized path
  planning on manifolds based on higher-dimensional continuation}.
\newblock \emph{The International Journal of Robotics Research}, 31\penalty0
  (2):\penalty0 201--215, 2012.

\bibitem[Porta et~al.(2014)Porta, Ros, Bohigas, Manubens, Rosales, and
  Jaillet]{Porta_RAM2014}
J.~M. Porta, L.~Ros, O.~Bohigas, M.~Manubens, C.~Rosales, and L.~Jaillet.
\newblock \href{http://doi.org/10.1109/MRA.2013.2287462}{The Cuik Suite:
  Analyzing the motion closed-chain multibody systems}.
\newblock \emph{IEEE Robotics and Automation Magazine}, 21\penalty0
  (3):\penalty0 105--114, 2014.

\bibitem[Posa et~al.(2016)Posa, Kuindersma, and Tedrake]{Posa_ICRA2016}
M.~Posa, S.~Kuindersma, and R.~Tedrake.
\newblock \href{http://doi.org/10.1109/ICRA.2016.7487270}{Optimization and
  stabilization of trajectories for constrained dynamical systems}.
\newblock In \emph{IEEE International Conference on Robotics and Automation},
  pages 1366--1373, 2016.

\bibitem[Potra and Yen(1991)]{Potra_JSM1991}
F.~A. Potra and J.~Yen.
\newblock \href{http://doi.org/10.1080/08905459108905138}{Implicit numerical
  integration for {Euler-Lagrange} equations via tangent space
  parametrization}.
\newblock \emph{Journal of Structural Mechanics}, 19\penalty0 (1):\penalty0
  77--98, 1991.

\bibitem[Reif(1979)]{Reif_ASFCS1979}
J.~H. Reif.
\newblock \href{http://doi.org/10.1109/SFCS.1979.10}{Complexity of the mover's
  problem and generalizations}.
\newblock In \emph{Annual Symposium on Foundations of Computer Science}, pages
  421--427, 1979.

\bibitem[Schulman et~al.(2014)Schulman, Y, Ho, Lee, Awwal, Bradlow, Pan, Patil,
  Goldberg, and Abbeel]{Schulman_IJRR2014}
J.~Schulman, Duan Y, J.~Ho, A.~Lee, I.~Awwal, H.~Bradlow, J.~Pan, S.~Patil,
  K.~Goldberg, and P.~Abbeel.
\newblock \href{http://doi.org/10.1177/0278364914528132}{Motion planning with
  sequential convex optimization and convex collision checking}.
\newblock \emph{The International Journal of Robotics Research}, 33\penalty0
  (9):\penalty0 1251--1270, 2014.

\bibitem[Shkolnik et~al.(2009)Shkolnik, Walter, and Tedrake]{Shkolnik_ICRA09}
A.~Shkolnik, M.~Walter, and R.~Tedrake.
\newblock \href{http://doi.org/10.1109/ROBOT.2009.5152874}{Reachability-guided
  sampling for planning under differential constraints}.
\newblock In \emph{IEEE International Conference on Robotics and Automation},
  pages 2859--2865, 2009.

\bibitem[Stilman(2007)]{Stilman_IROS07}
M.~Stilman.
\newblock \href{http://doi.org/10.1109/IROS.2007.4399305}{Task constrained
  motion planning in robot joint space}.
\newblock In \emph{IEEE/RSJ International Conference on Intelligent Robots and
  Systems}, pages 3074--3081, 2007.

\bibitem[{\c{S}}ucan and Kavraki(2012)]{Sucan_TRO2012}
I.~A. {\c{S}}ucan and L.~E. Kavraki.
\newblock \href{http://doi.org/10.1109/TRO.2011.2160466}{A sampling-based tree
  planner for systems with complex dynamics}.
\newblock \emph{IEEE Transactions on Robotics}, 28\penalty0 (1):\penalty0
  116--131, 2012.

\bibitem[{The Neuroscience and Robotics Lab, Northwestern
  University}(2016)]{TrepWeb}
{The Neuroscience and Robotics Lab, Northwestern University}.
\newblock Trep: Mechanical simulation and optimal control.
\newblock
  \href{http://murpheylab.github.io/trep}{http://murpheylab.github.io/trep},
  2016.

\bibitem[Zucker et~al.(2013)Zucker, Ratliff, Dragan, Pivtoraiko, Klingensmith,
  Dellin, Bagnell, and Srinivasa]{Zucker_IJRR2013}
M.~Zucker, N.~Ratliff, A.~D. Dragan, M.~Pivtoraiko, M.~Klingensmith, C.~M.
  Dellin, J.~A. Bagnell, and S.~S. Srinivasa.
\newblock \href{http://doi.org/10.1177/0278364913488805}{CHOMP: Covariant
  Hamiltonian optimization for motion planning}.
\newblock \emph{The International Journal of Robotics Research}, 32\penalty0
  (9-10):\penalty0 1164--1193, 2013.

\end{thebibliography}


\end{document}